\documentclass[11pt]{article}
\usepackage[english]{babel}
\usepackage[letterpaper,top=2.2cm,bottom=2.2cm,left=2.7cm,right=2.7cm]{geometry}
\usepackage[T1]{fontenc}
\usepackage{lmodern,microtype}
\usepackage{amsmath,amssymb,amsthm,amsfonts,mathtools}
\usepackage{bbm}
\usepackage{booktabs}
\usepackage{rotating}
\usepackage{graphicx}
\usepackage{xcolor}
\usepackage{csquotes}
\usepackage[backend=bibtex,style=alphabetic,natbib=true,sortcites=true,maxcitenames=2,maxbibnames=99,maxalphanames=6,minalphanames=3,isbn=false]{biblatex}
\DeclareFieldFormat[misc]{title}{\iffieldundef{eprint}{\mkbibemph{#1}}{\mkbibquote{#1\isdot}}} 
\usepackage{aliascnt}
\usepackage{hyperref}
\usepackage[capitalize,noabbrev]{cleveref}
\allowdisplaybreaks

\theoremstyle{plain}
\newtheorem{theorem}{Theorem}[section]
\newaliascnt{proposition}{theorem}

\aliascntresetthe{proposition}
\newaliascnt{lemma}{theorem}
\newtheorem{lemma}[lemma]{Lemma}
\aliascntresetthe{lemma}
\newaliascnt{corollary}{theorem}

\aliascntresetthe{corollary}
\newaliascnt{claim}{theorem}
\newtheorem{claim}[claim]{Claim}
\aliascntresetthe{claim}
\theoremstyle{definition}
\newaliascnt{definition}{theorem}
\newtheorem{definition}[definition]{Definition}
\aliascntresetthe{definition}
\theoremstyle{remark}
\newaliascnt{remark}{theorem}

\aliascntresetthe{remark}
\crefname{proposition}{Proposition}{Propositions}
\crefname{lemma}{Lemma}{Lemmas}
\crefname{corollary}{Corollary}{Corollaries}
\crefname{claim}{Claim}{Claims}
\crefname{definition}{Definition}{Definitions}
\crefname{remark}{Remark}{Remarks}
\crefformat{equation}{(#2#1#3)}
\Crefformat{equation}{Equation~(#2#1#3)}
\crefrangeformat{equation}{(#3#1#4) to~(#5#2#6)}
\crefmultiformat{equation}{(#2#1#3)}{ and~(#2#1#3)}{, (#2#1#3)}{, and~(#2#1#3)}
\crefname{appendix}{Appendix}{Appendices}
\Crefname{appendix}{Appendix}{Appendices}

\newcommand{\reals}{\mathbb{R}}
\newcommand{\naturals}{\mathbb{N}}
\newcommand{\lvec}[1]{\reflectbox{$\vec{\reflectbox{$#1$}}$}}
\newcommand{\para}[1]{\paragraph{#1}}
\newcommand{\eps}{\varepsilon}
\newcommand{\out}{y}
\newcommand{\outRV}{Y}
\newcommand{\outDom}{\mathcal{Y}}
\newcommand{\alg}{M}
\newcommand{\domain}{\mathcal{X}}
\newcommand{\dataset}{\boldsymbol{s}}
\newcommand{\view}{\boldsymbol{v}}
\newcommand{\prob}[2]{\ifx&#1&\mathbb{P}\else\underset{#1}{\mathbb{P}}\fi\left(#2\right)}
\newcommand{\expect}[2]{\ifx&#1&\mathbb{E}\else\underset{#1}{\mathbb{E}}\fi\left[#2\right]}
\newcommand{\loss}{\ell}
\newcommand{\lossFunc}[3]{\loss\left(#1; #2, #3 \right)}
\newcommand{\HockeyStick}[1]{\boldsymbol{H}_{#1}}
\newcommand{\HockeyStickDiv}[3]{\HockeyStick{#1}\left(#2\,\middle\Vert\,#3\right)}
\newcommand{\shuf}[2]{\mathcal{S}_{#1}\left(#2\right)}
\newcommand{\shufFunc}[3]{\mathcal{S}_{#1}\left(#2; #3\right)}
\newcommand{\alloc}[2]{\mathcal{A}_{#1}\left(#2\right)}
\newcommand{\add}[1]{\lvec{#1}}
\newcommand{\rem}[1]{\vec{#1}}
\newcommand{\privProf}[1]{\delta_{#1}}
\newcommand{\privProfAdd}[1]{\add{\delta}_{#1}}
\newcommand{\privProfRem}[1]{\rem{\delta}_{#1}}
\newcommand{\rep}[1]{\overset{\mathclap{\scriptscriptstyle\leftrightarrow}}{#1}}
\newcommand{\privProfRep}[1]{\rep{\delta}_{#1}}
\newcommand{\TV}{\operatorname{TV}}
\newcommand{\Bin}{\operatorname{Bin}}
\newcommand{\Phibar}{\overline{\Phi}}
\newcommand{\Pbar}[1]{\bar{P}_{#1}}
\newcommand{\Nrm}{\mathcal{N}}
\newcommand{\GaussMech}[1]{N_{#1}}
\newcommand{\Pmix}{P_{\mathrm{mix}}}
\newcommand{\Qmix}{Q_{\mathrm{mix}}}
\newenvironment{proofof}[1]{\begin{proof}[\normalfont\bfseries Proof of #1]\emergencystretch=2em}{\end{proof}}

\newcommand{\resFigOneUpperLowerMax}{1.3}
\newcommand{\resAllocGapHigh}{10}
\newcommand{\resAllocRatioEight}{1.5}
\newcommand{\resAllocRatioFour}{2}
\newcommand{\resAllocRatioLow}{5}
\newcommand{\resCloneProbHigh}{0.95}
\newcommand{\resPoissonGap}{10}
\newcommand{\resPoissonGapHigh}{1}
\newcommand{\resMaxStatSigmaOne}{3.7}
\newcommand{\resUpperSigmaOne}{4.5}
\newcommand{\resLRGainTwo}{15}
\newcommand{\resLRGainEight}{4}
\newcommand{\resUpperLRMax}{2.3}
\newcommand{\resUpperLRLow}{1.4}
\newcommand{\resGroupNRTwo}{7}
\newcommand{\resClonesGain}{7}
\newcommand{\resClonesCapSigma}{2}
\newcommand{\resCloneGridChange}{0.06}
\newcommand{\resCellsChange}{0.002}
\newcommand{\resPureFMTGainTwo}{3}
\newcommand{\resRRFMTWithin}{1}
\newcommand{\resLaplaceFMTAbove}{7}
\newcommand{\resGenericBelowMax}{0.75}
\newcommand{\resGenericAboveMin}{1}
\newcommand{\resGenericAboveMax}{3}
\newcommand{\resGenericAbovePct}{20}
\newcommand{\resHundredRRCross}{1.8}
\newcommand{\resGenericGroupCapped}{2}
\newcommand{\resShuffledSettings}{14}
\newcommand{\resMaxExamplesMillions}{1.3}
\newcommand{\resMinReportedEps}{0.83}
\newcommand{\resMaxReportedEps}{71}
\newcommand{\resSufNecMax}{1.39}
\newcommand{\resSufNecMedian}{1.19}
\newcommand{\resSufNecMaxAbout}{1.4}
\newcommand{\resSufNecMedianAbout}{1.2}
\newcommand{\resSufNecMaxLargeEps}{1.22}
\newcommand{\resUnampSetting}{20}
\newcommand{\resUnampPct}{0.4}
\newcommand{\resEnvelopeMedian}{20}
\newcommand{\resEnvelopeMax}{29}
\newcommand{\resEnvelopeMaxAbout}{30}
\newcommand{\resSingleCloneSufNecMax}{1.67}
\newcommand{\resSingleCloneSufNecMedian}{1.53}
\newcommand{\resNecGainCount}{19}
\newcommand{\resNecNoGainSetting}{20}
\newcommand{\resNecGainSetting}{2}
\newcommand{\resNecGainFrom}{2.77}
\newcommand{\resNecGainTo}{3.18}
\newcommand{\resNecMatchCount}{18}
\newcommand{\resNecMatchMin}{1.47}
\newcommand{\resNecMatchMax}{1.94}
\newcommand{\resNecMatchMinAbout}{1.5}
\newcommand{\resNecMatchMaxAbout}{2}
\newcommand{\resNecMatchLowFirstID}{1}
\newcommand{\resNecMatchLowFirst}{1.31}
\newcommand{\resNecMatchLowSecondID}{2}
\newcommand{\resNecMatchLowSecond}{1.17}
\newcommand{\resSufMatchMin}{1.52}
\newcommand{\resSufMatchMax}{2.41}
\newcommand{\resSufMatchSecond}{2.18}
\newcommand{\resMatchChange}{2.4}
\newcommand{\resCloneGridDpsgd}{0.7}

\title{Privacy Amplification by Shuffling\\ via Random Allocation among Clones}
\author{
    Vitaly Feldman\\
    Apple
    \and
    Moshe Shenfeld\footnote{Work partially done while author was an intern at Apple.}\\
    The Hebrew University of Jerusalem
}
\date{}

\begin{document}
\maketitle

\begin{abstract}
Shuffling is the batch sampling scheme used in most practical implementations of differentially private stochastic gradient descent (DP-SGD). Privacy amplification by shuffling is also the basis of the shuffle model of differential privacy, which is deployed in industry. Yet the privacy analysis of shuffling lags behind that of Poisson subsampling, the scheme assumed in the standard analysis of DP-SGD. For Gaussian noise addition, the known bounds for shuffling lose a factor of $\sqrt{\ln(n/\delta)}$ in $\eps$ relative to Poisson subsampling, and no numerically accurate accountant is available.

In this work we give a new general analysis of shuffling based on a reduction to $1$-out-of-$m$ random allocation, a sampling scheme for which nearly tight and efficiently computable privacy bounds were recently established. Here $m\approx e^{-\eps_0}n$, where $n$ is the number of shuffled elements and $\eps_0$ is the privacy parameter of each step for $\delta_0\approx\delta/n$. For the Gaussian mechanism with noise scale $\sigma=\Omega(\sqrt{\ln(n/\delta)})$ and $\ln(n/\delta)=O(\sqrt n)$, the reduction gives the asymptotically optimal $\eps=O\bigl(\sqrt{\ln(1/\delta)}/(\sigma\sqrt n)\bigr)$. Further, as the noise level grows, the bound approaches that of random allocation up to a factor $1+o(1)$, provided that $\delta$ decays sufficiently fast. The reduction also gives a dominating privacy loss distribution for shuffling, which composes across multiple passes with independent permutations and yields an accurate privacy accountant. At moderate and high noise the resulting bounds are close to those of random allocation. For the low-noise regime, where shuffling is substantially less private, we establish tighter lower bounds on an explicit shuffled Gaussian instance and show that our upper bounds closely track them.
\end{abstract}

\section{Introduction}
Privacy amplification by sampling is one of the central tools in the analysis of differentially private (DP) algorithms. A DP algorithm (or a sequence of DP algorithms) is run on randomly chosen elements of the dataset without revealing which elements were used, and this additional uncertainty improves the privacy guarantee \citep{KLNRS11}. The most important application of amplification by sampling is the differentially private stochastic gradient descent (DP-SGD) algorithm \citep{BST14, ACGMMT16}, in which gradients are computed on randomly chosen batches and privatized by clipping and Gaussian noise addition. The privacy analysis of DP-SGD assumes \emph{Poisson subsampling}, in which every element is included in every batch independently with the same probability. Independence makes the algorithm an adaptive composition of independently subsampled Gaussian mechanisms. Such a composition can be analyzed asymptotically and, more importantly for applications, accounted for numerically with high accuracy via the privacy loss distribution (PLD) \citep{MM18,SMM19,KJH20,GLW21,Google20}.

However, Poisson subsampling is rarely used in practical implementations, as independently resampling every batch is inefficient, hard to fit into standard training pipelines and tends to reduce the accuracy of the output model. In practice the data is instead \emph{shuffled}, that is, randomly permuted and then split into consecutive batches, while the privacy analysis still assumes Poisson subsampling \citep{PHKXDMVCT23,MHSBCCGKKL25}. Shuffling is also the basis of the privacy guarantees in the \emph{shuffle model} of differential privacy, in which every user applies a differentially private local randomizer to their data and a shuffler randomly permutes the reports before they are released \citep{BEMMRLRKTS17,EFMRTT19,CSUZZ19}. Privacy amplification by shuffling has been analyzed in a long line of work \citep{EFMRTT19,CSUZZ19,BBGN19,BKMT20,FMT21,FMT23,GDDKS21,GDDSK21,KHH22}. These analyses are nearly tight for pure DP algorithms, but for Gaussian noise addition they proceed through an $(\eps_0,\delta_0)$-DP guarantee of each step with $\delta_0\approx\delta/n$, where $n$ is the number of shuffled elements. Relative to Poisson subsampling this costs an additional $\sqrt{\ln(n/\delta)}$ asymptotic factor in $\eps$, and the numerical bounds are far from those of Poisson subsampling as well. The resulting discrepancy between the implementation of DP-SGD and its privacy analysis is well recognized, and recent work shows that in some regimes shuffling can be genuinely less private than Poisson subsampling \citep{CGKKMSZ24a,CGKKMSZ24b,ABDCH26}.

To address this discrepancy while maintaining the utility benefits of shuffling, the \emph{$1$-out-of-$t$ random allocation} sampling scheme was recently proposed in~\citep{CGHLKKMSZ24} (under the name \emph{balls-and-bins}). In this scheme, each element is assigned to a single step chosen uniformly at random out of $t$ steps, independently of the other elements. \citet{FS25ra,DCO25} extended it to $k$-out-of-$t$ random allocation, in which each element is assigned to $k$ steps chosen uniformly at random without replacement. Note that a shuffled pass of $t$ batches over a dataset of $n$ elements places each element in a single uniformly random batch, so the marginal distribution of the step in which an element participates is the same as in $1$-out-of-$t$ random allocation. Recent work has largely closed the gap between the privacy analysis of random allocation and that of Poisson subsampling, giving reductions to Poisson subsampling and bounds on R\'enyi divergence \citep{FS25ra,DCO25,SK26}, efficient and accurate numerical computation of the PLD \citep{FS26pld,FS26msfft}, and explicit and nearly tight asymptotic bounds \citep{FS26clt}. Yet in shuffling the step choices of different elements are coupled rather than independent, so existing techniques for random allocation do not apply. It has therefore remained unclear how the privacy of shuffling relates to that of random allocation, and in particular whether the extra $\sqrt{\ln(n/\delta)}$ factor in existing bounds for Gaussian shuffling can be removed.

\subsection{Our contributions}\label{sec:contrib}
\begin{figure}[t]
 \centering
 \includegraphics[width=0.66\linewidth]{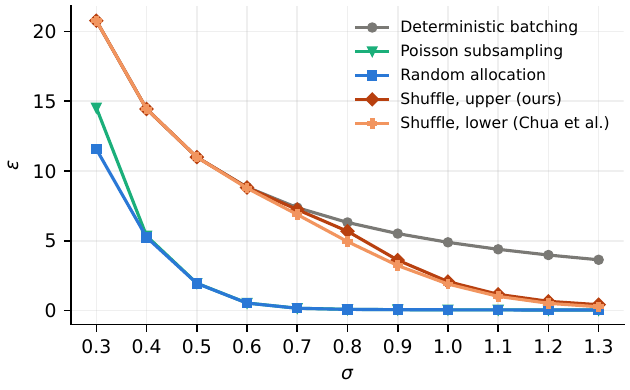}
 \caption{$\eps$ of one epoch of $t=10{,}000$ steps with batch size one, for $\delta=10^{-6}$, unit norm bound, and zero-out adjacency, as in \citep[Figure~1]{CGKKMSZ24a}. Poisson subsampling uses sampling probability $1/t$, our upper bound is the PLD envelope of the count-mixture bound in \cref{thm:reduction}, and the lower bound is from~\citep{CGKKMSZ24a}. The clones bound~\citep{FMT21} is not shown, since in this regime it is not lower than deterministic batching.}
 \label{fig:main}
\end{figure}

We establish a direct connection between shuffling and random allocation through a simple general reduction, which also yields an accurate PLD-based accountant. For the Gaussian mechanism, our reduction removes the $\sqrt{\ln(n/\delta)}$ factor and matches the asymptotics of random allocation and Poisson subsampling at moderate and high noise. At lower noise (including most published multi-epoch DP-SGD settings), where shuffling is known to be less private than Poisson subsampling, our numerical bounds closely track lower bounds for shuffling, as illustrated in \cref{fig:main}.

\para{A reduction from shuffling to random allocation.}
In our main result we show that the privacy of shuffling $n$ elements is upper bounded by that of $1$-out-of-$m$ random allocation of a dominating pair of a single step, where $m\approx e^{-\eps_0}n$ for a privacy parameter $\eps_0$ of each step, up to additive terms in $\delta$. The reduction applies to an arbitrary adaptive algorithm, and it compares a dataset containing a distinguished element $x$ with the dataset in which $x$ is replaced by a null element $\bot$ (zero-out adjacency, standard in DP-SGD accounting). Its only assumption is a one-sided privacy condition on the algorithm: for every element, its output distribution must be at least $e^{-\eps_0}$ times its output distribution on $\bot$, which we refer to as the \emph{reference distribution}, up to a small missing mass. For example, this holds when the algorithm is $(\eps_0,\delta_0)$-DP for $\delta_0\approx\delta/n$.

More formally, let $\alg(x,\view)$ be an algorithm that is applied to the elements in a uniformly random order and may depend arbitrarily on the view $\view$ of all previous responses, and let $Q_{\view}$ be its response distribution on $\bot$. We assume that
\[
 \HockeyStickDiv{e^{\eps_0}}{Q_{\view}}{\alg(x,\view)}\le\delta_0
 \qquad\text{for all } x,\view,
\]
where $\HockeyStick{\kappa}$ is the hockey-stick divergence. That is, we only require that the algorithm is $(\eps_0,\delta_0)$-DP in the add direction, in which $\bot$ is compared with $x$. We show that if a pair of distributions $(P,Q)$ dominates $\alg$~\citep{ZDW22} (meaning that $(P,Q)$ has at least as large a hockey-stick divergence as $(\alg(x,\view),Q_{\view})$ for all $x,\view$), then for $K\sim1+\Bin(n-1,e^{-\eps_0})$,
\begin{equation}\label{eq:intro-count-mixture}
 \privProf{\shuf{n}{\alg}}(\eps)
 \le\expect{K}{\privProf{\alloc{K}{P,Q}}(\eps)}+(1+e^{\eps})\,n e^{-\eps_0}\delta_0.
\end{equation}
Here $\privProf{\shuf{n}{\alg}}(\eps)$ denotes the privacy profile of the shuffled algorithm $\shuf{n}{\alg}$ on $n$ elements (the smallest $\delta$ for which it is $(\eps,\delta)$-DP), and $\alloc{k}{P,Q}$ denotes $1$-out-of-$k$ random allocation of an algorithm whose output distributions are $P$ on the distinguished element and $Q$ on $\bot$. We refer to this bound, which holds separately for the remove and the add directions, as the \emph{count-mixture bound}. The concentration of $K$ around $e^{-\eps_0}n$ and the fact that $\privProf{\alloc{k}{P,Q}}(\eps)$ is nonincreasing in $k$ imply a \emph{single-count bound} $\privProf{\alloc{m}{P,Q}}(\eps)$ at $m\approx e^{-\eps_0}n$ up to a small tail probability, which suffices for our asymptotic bounds (see \cref{thm:reduction} for both forms).

Our proof builds on the privacy blanket and clone decompositions in~\citep{BBGN19,FMT21}, which split each element's response distribution into a common component and a residual. Suppose first that the response of every common element (an element other than $x$) is a mixture that, with probability $e^{-\eps_0}$, outputs a fresh sample from the reference distribution $Q_{\view}$, which following~\citep{FMT21} we refer to as a \emph{clone}. Since every common element produces a clone independently with probability $e^{-\eps_0}$, the number of clones is distributed as $\Bin(n-1,e^{-\eps_0})$. We then reveal which common elements \emph{did not} produce a clone, together with their positions. Revealing this information can only help distinguish the two datasets, yet the distinguished response remains hidden at a uniformly random position among the clones, reducing the unrevealed part of the output to $1$-out-of-$K$ random allocation with $K-1$ clones. When the reference component holds only approximately ($\delta_0>0$), we fill the small deficit where a common element's density falls below $e^{-\eps_0}Q_{\view}$. This contributes an additive cost of $(1+e^{\eps})\,n e^{-\eps_0}\delta_0$ in $\delta$ to the privacy profile.

The key benefit of our reduction is that it allows us to use known bounds and accounting tools for random allocation, including the efficient computation of the PLD \citep{FS26pld,FS26msfft} and the Gaussian approximation of its privacy loss \citep{FS26clt}. Most notably, the mixture structure behind eq.~\cref{eq:intro-count-mixture} gives a way to compute a dominating pair and PLD for shuffling directly from the PLDs of random allocation (\cref{sec:pld-accounting}). We further combine the pairs obtained at many clone probabilities into a single tighter envelope pair of distributions. This removes the need to optimize over the choice of the clone probability and gives better bounds for composition of shuffled passes.

Our zero-out bounds can be used to derive two incomparable bounds for replacement adjacency (in which $x$ is replaced by another element $x'$ rather than by $\bot$; \cref{app:replacement}). The \emph{group bound} (\cref{lem:group}) combines the remove and add zero-out profiles via group privacy at roughly a factor-of-$2$ cost in $\eps$. The second \emph{neighbor-reference bound} (\cref{lem:neighbor-reference}) uses the response of one of the two neighboring elements as the reference distribution together with the replacement guarantee of the algorithm.

\para{Tight asymptotics for the Gaussian mechanism.}
With noise scale $\sigma$ and unit norm bound, the reference distribution is the noise distribution $\Nrm(0,\sigma^2I)$, and the assumption of the reduction holds with $\delta_0\approx\delta/n$ at $\eps_0=O\bigl(\sqrt{\ln(n/\delta)}/\sigma\bigr)$. Thus, whenever $\sigma=\Omega(\sqrt{\ln(n/\delta)})$ we have $\eps_0=O(1)$, so a constant fraction of the responses are clones. Combining the reduction with recent bounds for Gaussian random allocation \citep{FS26clt}, we show (\cref{thm:gauss-shuffle}) that for $\sigma=\Omega(\sqrt{\ln(n/\delta)})$ and $\ln(n/\delta)=O(\sqrt n)$, a shuffled pass over $n$ elements is $(\eps,\delta)$-DP under zero-out adjacency with $\eps=O\bigl(\sqrt{\ln(1/\delta)}/(\sigma\sqrt n)\bigr)$. This improves on the best previous bounds for shuffling with Gaussian noise by a factor of $\sqrt{\ln(n/\delta)}$, as those bounds are of order $\sqrt{\ln(n/\delta)\ln(1/\delta)}/(\sigma\sqrt n)$ \citep{FMT21,FMT23}. Overall, our bound matches the rate of random allocation and of Poisson subsampling with sampling probability $1/n$, and as the noise grows faster than $\sqrt{\ln(1/\delta)}$ (when $\ln(1/\delta)$ grows faster than $\ln n$ and slower than $\sqrt n$), the clone fraction approaches $1$ and the bound matches both up to a $1+o(1)$ factor.

\para{Tighter lower bounds.}
To complement our upper bounds, we derive tighter lower bounds for the one-dimensional Gaussian instance in~\citep{CGKKMSZ24a} (conjectured to be the worst case for shuffling), which we call the \emph{adversarial instance}, where common elements contribute $-1$ and the distinguished element contributes $1$ when present and $0$ when absent. Prior lower bounds on this instance threshold only the largest released response \citep{CGKKMSZ24a,CGKKMSZ24b}. While this is nearly optimal at very low noise, it does not capture the privacy loss at moderate noise. In \cref{sec:lower-bound} we instead analyze the optimal likelihood-ratio test directly: because all common responses are identically distributed, this test simplifies to thresholding a sum of independent scalar terms (\cref{lem:identical-ratio}), yielding a tight single-pass lower bound. To extend this bound to multiple epochs, we fold a collection of such tests into a discrete pair dominated by one shuffled pass (\cref{lem:pool}), which gives a lower bound on the PLD of this instance that composes directly across epochs. We evaluate the whole collection by making the summands nonnegative and then adapting the multiscale FFT convolution algorithm in~\citep{FS26msfft} to this problem. 

\subsection{Numerical evaluation}
The count-mixture bound (\cref{thm:reduction}) and PLD construction (\cref{sec:pld-accounting}) yield a numerical accountant that we evaluate in \cref{sec:evaluation}, together with our lower bounds.

\para{Gaussian mechanism.}
In the setting in~\citep{CGKKMSZ24a}, shown in \cref{fig:main}, our upper bound is within a factor of $\resFigOneUpperLowerMax$ of their lower bound for shuffling for every $\sigma\le1.2$. We also compare shuffling with random allocation and Poisson subsampling at $t=n=1000$ steps and $\delta=10^{-6}$ (\cref{fig:comparison}). At $\sigma\ge32$ the shuffling bound is within about $\resAllocGapHigh\%$ of the random allocation bound, at $\sigma=8$ it is larger by a factor of about $\resAllocRatioEight$, and at $\sigma\le2$ by a factor of about $\resAllocRatioLow$ or more.

A large part of this low-noise gap is a property of shuffling rather than of our analysis: intuitively, at low noise the $n-1$ common elements can make the distinguished response much easier to locate than it is among the null-element responses of random allocation. On the adversarial instance described above, our likelihood-ratio lower bound (\cref{sec:lower-bound}) is within a factor of $\resUpperLRMax$ of our upper bound for all $\sigma\le2$ and exceeds the upper bound for random allocation for all $\sigma\le16$.

\para{Replacement adjacency and pure DP.}
Under replacement adjacency, our bound for the Gaussian mechanism improves on the clones bound in~\citep{FMT21} by a factor of at least $\resClonesGain$ for $2\le\sigma\le128$ at $n=1000$ (\cref{fig:gauss-replacement}). For pure DP algorithms (\cref{fig:pure-clones}), our replacement bounds for randomized response and the Laplace mechanism improve on those in~\citep{FMT21} and nearly match the algorithm-specific bounds in~\citep{FMT23}, whereas for a generic pure DP algorithm our bound improves on that in~\citep{FMT21} only for some values of $\eps_0$.

\para{Applications to DP-SGD.}
Finally, in \cref{sec:dpsgd} we evaluate the tightness of our bound in $20$ multi-epoch DP-SGD settings taken from published papers (in $\resShuffledSettings$ of them the released code trains with shuffled or fixed batches, while the paper reports a guarantee for Poisson subsampling). In every setting, the noise that our upper bound shows to be sufficient for the reported privacy guarantee is at most about $\resSufNecMaxAbout$ times (with a median of $\resSufNecMedianAbout$) the noise that our multi-pass lower bound for shuffling (\cref{sec:lower-bound}) shows to be necessary. In $\resNecMatchCount$ of these settings, this lower bound also shows that shuffling needs about $\resNecMatchMinAbout$ to $\resNecMatchMaxAbout$ times the noise of Poisson subsampling.

\subsection{Related work}\label{sec:related}
\para{Shuffle model.}
The shuffle model was proposed by \citet{BEMMRLRKTS17}, and the formal analysis of privacy amplification by shuffling was initiated by \citet{EFMRTT19} and \citet{CSUZZ19}. \citet{BBGN19} introduced the privacy blanket decomposition, and \citet{FMT21} gave a nearly optimal analysis for pure DP algorithms by hiding the distinguished response among clones of the two neighboring inputs. \citet{FMT23} and \citeauthor{GDDKS21}~\citep{GDDKS21,GDDSK21} extended these analyses to approximate and R\'enyi DP, and \citet{KHH22} developed numerical accounting for shuffling of algorithms with discrete outputs. All of these bounds are based on a pure DP or $(\eps_0,\delta_0)$-DP parameter of the algorithm, and for the Gaussian mechanism they inherit the overhead discussed above. \citet{TL26}\footnote{The datasets $(0,\ldots,0,1)$ and $(0,\ldots,0)$ that \citet{LT22} analyze for the shuffled Gaussian mechanism are not a dominating pair for shuffling.} avoid this parameter by analyzing the privacy blanket bound~\citep{BBGN19} directly, under replacement adjacency. They characterize the bound asymptotically in $n$ and numerically evaluate it for a one-dimensional Gaussian mechanism, but the resulting amplification decreases exponentially in the square root of the dimension \citep[Remark~3.5]{TL26}. Instead, we use the response of the null element as the reference distribution, and for the Gaussian mechanism both the clone probability and the dominating pair are the same in every dimension.

In the other direction, \citet{CGKKMSZ24a,CGKKMSZ24b} showed analytically that shuffled batch sampling with Gaussian noise can be substantially less private than Poisson subsampling, and \citet{ABDCH26} demonstrated this gap by auditing shuffled DP-SGD. The lower bound in~\citep{CGKKMSZ24a,CGKKMSZ24b} uses the instance and largest-response test described in \cref{sec:contrib}, and serves as a baseline in \cref{fig:main,fig:comparison,fig:dpsgd}.

\para{Random allocation.}
\citet{CGHLKKMSZ24} and \citet{CCGHST25} proposed balls-and-bins sampling for DP-SGD and for matrix mechanisms, respectively, and estimated its privacy parameters by Monte Carlo simulation (an estimate later improved in~\citep{DG26}). Analytical bounds for random allocation were obtained through a reduction to Poisson subsampling and two direct analyses \citep{FS25ra}, as well as through R\'enyi divergence bounds~\citep{DCO25,SK26}. Subsequent work developed numerical PLD accounting for random allocation \citep{FS26pld,FS26msfft}, analyzed its privacy loss via Gaussian approximation \citep{FS26clt}, and compared the sampling schemes asymptotically \citep{DO26}. Finally, random check-ins \citep{BKMT20} contain $1$-out-of-$t$ random allocation as a special case, but their analysis relies on shuffling.

\section{Preliminaries}\label{sec:preliminaries}
For $n\in\naturals$ we write $[n]\coloneqq\{1,\ldots,n\}$. For a distribution $P$ and a point $\omega$ in its sample space, we write $P(\omega)$ for the probability of $\omega$, or its density with respect to a fixed dominating measure in the continuous case. We write $[x]_+\coloneqq\max\{x,0\}$, and all logarithms are natural. The proof of \cref{lem:hs-facts} appears in \cref{app:reduction-proofs}.

\subsection{Algorithms, adjacency and sampling schemes}
Let $\domain$ be a data domain, let $\bot\notin\domain$ be a \emph{null element}, and write $\domain'\coloneqq\domain\cup\{\bot\}$. Let $\outDom$ be an output space. An adaptive algorithm $\alg$ maps each element $x\in\domain'$ and finite sequence $\view$ of previous outputs in $\outDom$ to a distribution $\alg(x,\view)$ over $\outDom$ (in particular, $\alg$ may also depend on the step index). We refer to a sample from $\alg(x,\view)$ as the \emph{response} of $x$, and to $Q_{\view}\coloneqq\alg(\bot,\view)$ as the \emph{reference distribution} at view $\view$.

Given a dataset $\dataset=(x_1,\ldots,x_n)\in(\domain')^n$, the \emph{shuffled algorithm} $\shuf{n}{\alg}$ chooses a uniformly random permutation $\pi$ of $[n]$ and generates $\out_1,\ldots,\out_n$ sequentially by drawing $\out_i\sim\alg(x_{\pi(i)},\view_{i-1})$ for $i=1,\ldots,n$, where $\view_{i-1}\coloneqq(\out_1,\ldots,\out_{i-1})$ is the \emph{view} before step $i$, and outputs $(\out_1,\ldots,\out_n)$. We write $\shufFunc{n}{\alg}{\dataset}$ for the resulting output distribution over $\outDom^n$. This models a single shuffled pass of an iterative algorithm such as DP-SGD with batches of size one.\footnote{Shuffling $n$ batches of size $B$ reduces to shuffling $n$ elements with batch size $1$ by revealing the $B-1$ other elements in each batch (\cref{app:dpsgd}).} In DP-SGD the response at each step is the noisy gradient, and the model parameters are updated after each step. The shuffled algorithm also captures the shuffle model, where $\alg$ applies a \emph{local randomizer} \citep{EFMRTT19} to each user's data and ignores the view.

Two datasets in $(\domain')^n$ are \emph{adjacent} under \emph{zero-out adjacency} \citep{PHKXDMVCT23} if one is obtained from the other by replacing a single element $x\in\domain$ with $\bot$. Since the distribution of $\shufFunc{n}{\alg}{\cdot}$ is invariant to the order of the input dataset, we write two adjacent datasets as $(\dataset,x)$ and $(\dataset,\bot)$, where $\dataset=(x_1,\ldots,x_{n-1})\in(\domain')^{n-1}$ collects the $n-1$ \emph{common elements} and $x$ (or $\bot$ in $(\dataset,\bot)$) is the \emph{distinguished element}. Following \citep{FS25ra,FS26pld}, we refer to comparing $(\dataset,x)$ against $(\dataset,\bot)$, with $(\dataset,x)$ as the first argument, as the \emph{remove direction}, and to the reverse comparison as the \emph{add direction}. We treat \emph{replacement adjacency}, in which $x$ is replaced by another element $x'\in\domain$, in \cref{app:replacement}.

In \emph{$1$-out-of-$t$ random allocation}, the distinguished element is assigned to a single step chosen uniformly at random from $[t]$, while all other steps use the null element. For our reduction it suffices to consider this scheme for a fixed pair of distributions $(P,Q)$ over a common space, which represent the responses of $x$ and $\bot$. We denote the resulting $t$-step algorithm by $\alloc{t}{P,Q}$. Its output distributions on $x$ and $\bot$, which we refer to as the \emph{allocation pair}, are
\begin{equation}\label{eq:alloc-pair}
 \Pbar{t}\coloneqq\frac1t\sum_{j=1}^{t}Q^{j-1}\times P\times Q^{t-j}
 \qquad\text{and}\qquad
 Q^{t},
\end{equation}
where $Q^{t}$ denotes the product of $t$ independent copies of $Q$. Similarly, in $t$-step \emph{Poisson subsampling} of $(P,Q)$ with sampling probability $\lambda$, the distinguished element is included at each step independently with probability $\lambda$, giving output distributions $((1-\lambda)Q+\lambda P)^t$ on $x$ and $Q^t$ on $\bot$.

\subsection{Privacy notions}
We measure privacy through the hockey-stick divergence and the privacy loss distribution, which characterize $(\eps,\delta)$-DP and allow tight numerical accounting.

\begin{definition}[{Hockey-stick divergence and privacy loss \citep{BKOZB12,DR16}}]\label{def:hs}
For distributions $P,Q$ over a common space, the \emph{privacy loss} of a point $\omega$ in this space is $\lossFunc{\omega}{P}{Q}\coloneqq\ln(P(\omega)/Q(\omega))$ (set to $+\infty$ when $Q(\omega)=0<P(\omega)$ and $-\infty$ when $P(\omega)=0<Q(\omega)$, with $e^{-\infty}=0$), and the \emph{privacy loss distribution (PLD)} is the distribution of $\lossFunc{\omega}{P}{Q}$ for $\omega\sim P$. For $\kappa\ge0$ the \emph{$\kappa$-hockey-stick divergence} is
\[
 \HockeyStickDiv{\kappa}{P}{Q}\coloneqq\int\left[P(\omega)-\kappa Q(\omega)\right]_+d\omega
 =\expect{\omega\sim P}{\left[1-\kappa e^{-\lossFunc{\omega}{P}{Q}}\right]_+}.
\]
\end{definition}

We will use four elementary properties of the hockey-stick divergence. 
\begin{lemma}\label{lem:hs-facts}
Given $\kappa\ge0$ and distributions $P,Q,P',Q'$ over a common space, the following hold.
\begin{enumerate}
 \item (Post-processing) For every randomized mapping $\varphi$, $\HockeyStickDiv{\kappa}{\varphi(P)}{\varphi(Q)}\le\HockeyStickDiv{\kappa}{P}{Q}$.
 \item (Convexity and monotonicity) $\HockeyStickDiv{\kappa}{P}{Q}\in[0,1]$, and $\kappa\mapsto\HockeyStickDiv{\kappa}{P}{Q}$ is convex and nonincreasing on $[0,\infty)$.
 \item (Triangle inequality) $\HockeyStickDiv{\kappa}{P}{Q}\le\HockeyStickDiv{\kappa}{P'}{Q'}+\TV(P,P')+\kappa\TV(Q,Q')$, where $\TV$ is the total variation distance.
 \item (Revealed mixtures) If $(w_i)_{i}$ are probability weights and, for every $i$, $P_i$ and $Q_i$ are distributions over a common space, then $\HockeyStickDiv{\kappa}{\sum_i w_i\,(\{i\}\times P_i)}{\sum_i w_i\,(\{i\}\times Q_i)}=\sum_i w_i\HockeyStickDiv{\kappa}{P_i}{Q_i}$.
\end{enumerate}
\end{lemma}

\begin{definition}[{Privacy profile \citep{BBG18}}]\label{def:profile}
For an algorithm $\alg$ defined on pairs of adjacent datasets and $\eps\ge0$, the \emph{remove} and \emph{add} privacy profiles are
\[
 \privProfRem{\alg}(\eps)\coloneqq\sup_{\dataset,x}\HockeyStickDiv{e^{\eps}}{\alg(\dataset,x)}{\alg(\dataset,\bot)},
 \qquad
 \privProfAdd{\alg}(\eps)\coloneqq\sup_{\dataset,x}\HockeyStickDiv{e^{\eps}}{\alg(\dataset,\bot)}{\alg(\dataset,x)},
\]
and the \emph{privacy profile} is $\privProf{\alg}(\eps)\coloneqq\max\{\privProfRem{\alg}(\eps),\privProfAdd{\alg}(\eps)\}$. For an adaptive algorithm $\alg(x,\view)$, the suprema run over $x\in\domain$ and the view $\view$, comparing $\alg(x,\view)$ with $Q_{\view}$. For $\eps\ge0$ and $\delta\in[0,1]$, we say that $\alg$ is \emph{$(\eps,\delta)$-differentially private (DP)} if $\privProf{\alg}(\eps)\le\delta$ \citep{DKMMN06}.
\end{definition}

In particular, by \cref{eq:alloc-pair}, the privacy profiles of $1$-out-of-$t$ random allocation of a pair $(P,Q)$ are
\begin{equation}\label{eq:alloc-profiles}
 \privProfRem{\alloc{t}{P,Q}}(\eps)=\HockeyStickDiv{e^{\eps}}{\Pbar{t}}{Q^t},
 \qquad
 \privProfAdd{\alloc{t}{P,Q}}(\eps)=\HockeyStickDiv{e^{\eps}}{Q^t}{\Pbar{t}}.
\end{equation}
The PLD of the allocation pair can be computed efficiently and accurately \citep{FS26pld,FS26msfft}, and we rely on this computation in \cref{sec:pld-accounting,sec:evaluation}.

\subsection{Dominating pairs}\label{sec:dom}
To analyze adaptive algorithms and compose them across passes, we summarize the worst-case privacy loss of an algorithm by a single fixed pair of distributions.

\begin{definition}[{Dominating pair \citep{ZDW22}}]\label{def:dom}
A pair of distributions $(P,Q)$ \emph{dominates} a pair $(P',Q')$ if $\HockeyStickDiv{\kappa}{P'}{Q'}\le\HockeyStickDiv{\kappa}{P}{Q}$ for all $\kappa\ge0$. We say that a pair $(P,Q)$ \emph{dominates} the algorithm $\alg$ (in the remove direction) if $(P,Q)$ dominates $(\alg(x,\view),Q_{\view})$ for every $x\in\domain$ and every view $\view$. For an algorithm on adjacent datasets, such as $\shuf{n}{\alg}$, domination in the remove and add directions refers to the pairs in \cref{def:profile}.
\end{definition}
A \emph{tightly dominating} pair (one with $\HockeyStickDiv{\kappa}{P}{Q}=\sup_{x,\view}\HockeyStickDiv{\kappa}{\alg(x,\view)}{Q_{\view}}$ for all $\kappa\ge0$) always exists \citep[Proposition~8]{ZDW22}, and therefore assuming domination in the remove direction is not a restriction. Our reduction relies on the following characterization of domination as post-processing.
\begin{lemma}[{\citep[Theorem~2.5]{KOV15}}]\label{lem:kov}
A pair of distributions $(P,Q)$ dominates the pair $(P',Q')$ if and only if there is a randomized mapping $\varphi$ with $\varphi(P)=P'$ and $\varphi(Q)=Q'$.
\end{lemma}
In particular, if a pair $(P,Q)$ dominates $\alg$, then for every $x,\view$ there is a randomized mapping $\varphi_{x,\view}$ with
\begin{equation}\label{eq:local-simulation}
 \varphi_{x,\view}(P)=\alg(x,\view),\qquad \varphi_{x,\view}(Q)=Q_{\view}.
\end{equation}
Thus every invocation of $\alg$ is a view-dependent post-processing of $(P,Q)$, and by symmetry of \cref{lem:kov}, $(Q,P)$ automatically dominates $\alg$ in the add direction.

We also use two standard properties of dominating pairs. First, if algorithms $\alg_1,\ldots,\alg_E$ are dominated by pairs $(P_1,Q_1),\ldots,(P_E,Q_E)$ (independent of the dataset), then their adaptive composition is dominated by $(P_1\times\cdots\times P_E, Q_1\times\cdots\times Q_E)$, whose PLD is the convolution of the individual PLDs \citep[Theorem~10]{ZDW22}. Second, if several pairs $(P_1,Q_1),\ldots,(P_r,Q_r)$ each dominate a pair $(P',Q')$, then by convexity of the hockey-stick divergence (\cref{lem:hs-facts}), $\kappa\mapsto\HockeyStickDiv{\kappa}{P'}{Q'}$ lies below the lower convex hull $H_*$ of $\kappa\mapsto\min_{i\in[r]}\HockeyStickDiv{\kappa}{P_i}{Q_i}$, and there is a single tighter dominating pair $(P_*,Q_*)$ whose hockey-stick divergence is $H_*$ \citep[Lemma~9]{ZDW22} (see also \citep{DGKKM22}). 

\subsection{The Gaussian mechanism}
The \emph{Gaussian mechanism} $\GaussMech{\sigma}$ with noise scale $\sigma$ and unit norm bound is the algorithm
\[
 \GaussMech{\sigma}(x,\view)=\Nrm(f_{\view}(x),\sigma^2I_d),\qquad \|f_{\view}(x)\|_2\le1,\qquad f_{\view}(\bot)=0,
\]
where $f_{\view}\colon\domain'\to\reals^d$ is an arbitrary function chosen adaptively based on the view, for example a clipped gradient at the current model. We leave the functions $f_{\view}$ implicit in the notation, and unless they are specified, a statement about $\GaussMech{\sigma}$ holds for every choice of them. The reference distribution is $Q_{\view}=\Nrm(0,\sigma^2I_d)$ and therefore the norm bound is a zero-out sensitivity bound. We refer to $(P,Q)=(\Nrm(1,\sigma^2),\Nrm(0,\sigma^2))$ as the \emph{Gaussian pair}.

\begin{claim}[{\citep[Lemma~3]{BW18}}]\label{clm:gauss-dom}
For every $\sigma>0$, the Gaussian pair $(\Nrm(1,\sigma^2),\Nrm(0,\sigma^2))$ dominates the Gaussian mechanism $\GaussMech{\sigma}$.
\end{claim}

The privacy profile of the Gaussian pair is \citep[Theorem~8]{BW18}
\begin{equation}\label{eq:gauss-profile}
 \HockeyStickDiv{e^{\eps}}{\Nrm(1,\sigma^2)}{\Nrm(0,\sigma^2)}
 =\Phi\left(\frac1{2\sigma}-\eps\sigma\right)-e^{\eps}\Phi\left(-\frac1{2\sigma}-\eps\sigma\right),
\end{equation}
where $\Phi$ is the standard normal distribution function. Since the Gaussian pair is symmetric under reflection, the same expression bounds both the remove and add privacy profiles of $\GaussMech{\sigma}$ (by \cref{clm:gauss-dom}).

\section{From shuffling to random allocation}\label{sec:reduction}
In this section we reduce the privacy analysis of shuffling to that of random allocation. We first state and prove the reduction in count-mixture and single-count forms (\cref{sec:statement}), and then build a dataset-independent dominating pair and PLD for shuffling (\cref{sec:pld-accounting}); replacement adjacency is treated in \cref{app:replacement}. Deferred proofs appear in \cref{app:reduction-proofs}.

\subsection{The reduction}\label{sec:statement}
Formalizing the clones argument outlined in \cref{sec:contrib} yields our main reduction.

\begin{theorem}[Shuffling to random allocation]\label{thm:reduction}
Given $n\in\naturals$, $\eps_0>0$, $\delta_0\in[0,1]$, and an algorithm $\alg$ dominated by a pair of distributions $(P,Q)$, if $\privProfAdd{\alg}(\eps_0)\le\delta_0$, then for every $\eps\ge0$,
\begin{equation}\label{eq:main-profile}
 \privProfRem{\shuf{n}{\alg}}(\eps)\le\sum_{k=1}^{n}w_k\,\privProfRem{\alloc{k}{P,Q}}(\eps)+(1+e^{\eps})\xi,
\end{equation}
where $p\coloneqq e^{-\eps_0}$, $w_k\coloneqq\prob{}{1+\Bin(n-1,p)=k}$ for $k\in[n]$, and $\xi\coloneqq1-(1-p\delta_0)^{n-1}$. In particular, for every $\eps\ge0$, $\beta\in(0,1)$, and integer $m\ge1$ with $\prob{}{\Bin(n-1,p)<m-1}\le\beta$,
\begin{equation}\label{eq:cutoff}
 \privProfRem{\shuf{n}{\alg}}(\eps)\le\privProfRem{\alloc{m}{P,Q}}(\eps)+\beta+(1+e^{\eps})(n-1)p\,\delta_0;
\end{equation}
this condition holds for $m=\max\Bigl\{1,\bigl\lfloor pn-\sqrt{2pn\ln(1/\beta)}\bigr\rfloor\Bigr\}$. Both bounds also hold with the add profiles $\privProfAdd{\shuf{n}{\alg}}$, $\privProfAdd{\alloc{k}{P,Q}}$, and $\privProfAdd{\alloc{m}{P,Q}}$ in place of the remove profiles.
\end{theorem}

We prove \cref{thm:reduction} in two steps. For $p\in(0,1)$, we say that a distribution $P'$ over $\outDom$ \emph{contains a $p$-fraction} of a distribution $Q'$ over $\outDom$ if $P'(\omega)\ge p\,Q'(\omega)$ for every $\omega\in\outDom$, that is, $P'=pQ'+(1-p)r$ for some distribution $r$ over $\outDom$ (equivalently, $\HockeyStickDiv{1/p}{Q'}{P'}=0$). First, when the response distribution of every common element already contains a $p$-fraction of the reference distribution, we show that the shuffled output distributions are dominated by a binomial mixture of the allocation pairs $(\Pbar{k},Q^k)$.

\begin{lemma}[Clones reduction]\label{lem:exact-mixture}
Given $n\in\naturals$, $p\in(0,1)$, an algorithm $\alg$, an element $x\in\domain$, $n-1$ elements $\dataset=(x_1,\ldots,x_{n-1})$ in the domain of $\alg$, and a pair of distributions $(P,Q)$, if for every view $\view$ the pair $(P,Q)$ dominates $(\alg(x,\view),Q_{\view})$ and $\alg(x_j,\view)$ contains a $p$-fraction of $Q_{\view}$ for every $j\in[n-1]$, then $\bigl(\shufFunc{n}{\alg}{(\dataset,x)},\shufFunc{n}{\alg}{(\dataset,\bot)}\bigr)$ is dominated by the \emph{mixture pair}
\begin{equation}\label{eq:mixture-pair}
 \Pmix\coloneqq\sum_{k=1}^{n}w_k\,\bigl(\{k\}\times\Pbar{k}\bigr),\qquad
 \Qmix\coloneqq\sum_{k=1}^{n}w_k\,\bigl(\{k\}\times Q^k\bigr),
\end{equation}
and the reversed pair is dominated by $(\Qmix,\Pmix)$, where $\Pbar{k}$ is defined in \cref{eq:alloc-pair} and $w_k\coloneqq\prob{}{1+\Bin(n-1,p)=k}$ for $k\in[n]$. In particular, if $(P,Q)$ dominates $\alg$ and $\privProfAdd{\alg}(\ln(1/p))=0$, then $(\Pmix,\Qmix)$ dominates $\shuf{n}{\alg}$ in the remove direction and $(\Qmix,\Pmix)$ dominates it in the add direction.
\end{lemma}
\begin{proof}
By assumption, for every $j\in[n-1]$ and every view $\view$ we can write $\alg(x_j,\view)=pQ_{\view}+(1-p)r_{j,\view}$ for some distribution $r_{j,\view}$ over $\outDom$. This gives the following way to run the shuffled algorithm on both datasets. Every common element $x_j$ flips an independent coin that equals $1$ with probability $p$, and when its turn comes at view $\view$ it responds with a fresh sample from $Q_{\view}$ (a clone) if its coin equals $1$ and with a sample from $r_{j,\view}$ otherwise. Let $R$ consist of the indices of the common elements whose coin equals $0$ together with their positions in the shuffled order, and let $K\coloneqq n-|R|$, so that $\prob{}{K=k}=w_k$. The coins and the permutation do not depend on the distinguished element, and thus $R$ has the same distribution on both datasets. Given $R$, the distinguished element and the $K-1$ clones occupy the other $K$ positions in a uniformly random order.

Fix a realization of $R$ with $K=k$, and let $i_1<\cdots<i_k$ be the positions that $R$ does not assign to common elements. By \cref{lem:kov}, for every view $\view$ there is a randomized mapping $\varphi_{x,\view}$ satisfying \cref{eq:local-simulation}. Define $\psi_R$ to be the randomized mapping that takes $z=(z_1,\ldots,z_k)$ in the sample space of $\Pbar{k}$ and $Q^k$ and generates the $n$ responses sequentially in the shuffled order: at each position assigned by $R$ to a common element $x_j$, it draws a response from $r_{j,\view}$ using the current view $\view$, and at each remaining position $i_l$ it outputs $\varphi_{x,\view}(z_l)$. Under $Q^k$ the coordinates of $z$ are independent, and under $\Pbar{k}$ they are independent given the uniformly random index $J\in[k]$ of the coordinate drawn from $P$. Since the view at position $i_l$ depends only on $z_1,\ldots,z_{l-1}$ and the internal randomness of $\psi_R$, the response at $i_l$ is distributed as $\alg(x,\view)$ when $z_l\sim P$ and as $Q_{\view}$ when $z_l\sim Q$. Since $i_J$ is uniform over $\{i_1,\ldots,i_k\}$, as is the position of the distinguished element given $R$, the mapping $\psi_R$ takes $\Pbar{k}$ and $Q^k$ to the conditional output distributions given $R$ on $(\dataset,x)$ and $(\dataset,\bot)$, respectively. Thus the mapping that, given $(k,z)$, draws $R$ from its conditional distribution given $K=k$ and then applies $\psi_R$ takes $\Pmix$ to $\shufFunc{n}{\alg}{(\dataset,x)}$ and $\Qmix$ to $\shufFunc{n}{\alg}{(\dataset,\bot)}$, and both domination claims follow from \cref{lem:kov}. For the last claim, suppose $(P,Q)$ dominates $\alg$ and $\privProfAdd{\alg}(\ln(1/p))=0$. Then for every view $\view$, $(P,Q)$ dominates $(\alg(x,\view),Q_{\view})$ for all $x\in\domain$, and $\alg(x',\view)$ contains a $p$-fraction of $Q_{\view}$ for all $x'\in\domain'$ (including $x'=\bot$, where $\alg(\bot,\view)=Q_{\view}$), so the hypotheses hold for every $x\in\domain$ and $\dataset\in(\domain')^{n-1}$.
\end{proof}

Under the assumption of \cref{thm:reduction}, the response distribution of a common element contains a $p$-fraction of $Q_{\view}$ only up to a small missing mass. We fill this mass by extending $\alg$ with a filled copy of each element and replacing only the common elements with their filled copies, keeping the response of the distinguished element unchanged. The following lemma bounds the resulting total variation cost.

\begin{lemma}[Cost of filling the common responses]\label{lem:extension}
Given $n\in\naturals$, $\eps_0>0$, $\delta_0\in[0,1]$, and an algorithm $\alg$, if $\privProfAdd{\alg}(\eps_0)\le\delta_0$, then $\alg$ can be extended by adding a \emph{filled copy} $\widetilde{x}'\notin\domain'$ of every $x'\in\domain'$ such that, for every view $\view$, $\alg(\widetilde{x}',\view)$ contains a $p$-fraction of $Q_{\view}$ and $\TV\bigl(\alg(x',\view),\alg(\widetilde{x}',\view)\bigr)\le p\delta_0$, where $p\coloneqq e^{-\eps_0}$. Moreover, for every $\dataset=(x_1,\ldots,x_{n-1})\in(\domain')^{n-1}$ and $x\in\domain$, the dataset $\widetilde{\dataset}\coloneqq(\widetilde{x}_1,\ldots,\widetilde{x}_{n-1})$ satisfies
\[
 \TV\bigl(\shufFunc{n}{\alg}{(\dataset,x)},\shufFunc{n}{\alg}{(\widetilde{\dataset},x)}\bigr)\le\xi
 \qquad\text{and}\qquad
 \TV\bigl(\shufFunc{n}{\alg}{(\dataset,\bot)},\shufFunc{n}{\alg}{(\widetilde{\dataset},\bot)}\bigr)\le\xi,
\]
where $\xi\coloneqq1-(1-p\delta_0)^{n-1}$.
\end{lemma}
\begin{proof}
For every $x'\in\domain'$ and view $\view$, we define $\alg(\widetilde{x}',\view)$ by applying \cref{lem:floor} (in \cref{app:reduction-proofs}) to $\alg(x',\view)$ and $Q_{\view}$ with $p=e^{-\eps_0}$, lifting $\alg(x',\view)$ up to $p\,Q_{\view}$ wherever it falls below $p\,Q_{\view}$ and scaling down its excess elsewhere. By \cref{lem:floor} and $\privProfAdd{\alg}(\eps_0)\le\delta_0$, $\alg(\widetilde{x}',\view)$ contains a $p$-fraction of $Q_{\view}$ and satisfies $\TV\bigl(\alg(x',\view),\alg(\widetilde{x}',\view)\bigr)\le p\delta_0$ (for $x'=\bot$ the distance is zero, since $\alg(\bot,\view)=Q_{\view}$).

We couple executions $(\outRV_1,\ldots,\outRV_n)$ on $(\dataset,x)$ and $(\widetilde\outRV_1,\ldots,\widetilde\outRV_n)$ on $(\widetilde{\dataset},x)$ under the same permutation $\pi$. Sequentially for $i=1,\ldots,n$, while both executions have the same view $\view$ before step $i$, we set $\widetilde\outRV_i=\outRV_i$ if position $i$ holds the distinguished element (both follow $\alg(x,\view)$). If it holds a common element ($x_j$ and $\widetilde{x}_j$), we couple $\alg(x_j,\view)$ and $\alg(\widetilde{x}_j,\view)$ to agree with probability at least $1-p\delta_0$; once the views differ, we continue independently. Since $n-1$ positions hold common elements, the two executions coincide with probability at least $(1-p\delta_0)^{n-1}$, yielding the first bound. The same coupling on $(\dataset,\bot)$ and $(\widetilde{\dataset},\bot)$ (where the distinguished element responds with $Q_{\view}$ in both runs) gives the second.
\end{proof}

\begin{proof}[Proof of \cref{thm:reduction}]
Fix $\dataset\in(\domain')^{n-1}$ and $x\in\domain$. Extend $\alg$ as in \cref{lem:extension}, let $\widetilde{\dataset}$ be the dataset of the filled copies of the elements of $\dataset$, and let $U_x$, $U_\bot$, $\widetilde U_x$ and $\widetilde U_\bot$ be the output distributions of $\shuf{n}{\alg}$ on $(\dataset,x)$, $(\dataset,\bot)$, $(\widetilde{\dataset},x)$ and $(\widetilde{\dataset},\bot)$, respectively. Since $\alg$ remains dominated by $(P,Q)$ on $\domain$ and its responses to the elements of $\widetilde{\dataset}$ contain a $p$-fraction of the reference distribution, \cref{lem:exact-mixture} shows that $(\Pmix,\Qmix)$ dominates $(\widetilde U_x,\widetilde U_\bot)$. The triangle inequality and the revealed-mixture property in \cref{lem:hs-facts}, together with \cref{lem:extension} and the allocation profiles in \cref{eq:alloc-profiles}, then give
\begin{align*}
 \HockeyStickDiv{e^{\eps}}{U_x}{U_\bot}
 &\le\HockeyStickDiv{e^{\eps}}{\widetilde U_x}{\widetilde U_\bot}+(1+e^{\eps})\xi\\
 &\le\HockeyStickDiv{e^{\eps}}{\Pmix}{\Qmix}+(1+e^{\eps})\xi
 =\sum_{k=1}^{n}w_k\,\privProfRem{\alloc{k}{P,Q}}(\eps)+(1+e^{\eps})\xi.
\end{align*}
Taking the supremum over $(\dataset,x)$ yields \cref{eq:main-profile}, and the add direction follows in the same way since $(\Qmix,\Pmix)$ dominates $(\widetilde U_\bot,\widetilde U_x)$.

To obtain \cref{eq:cutoff}, note that $(\Pbar{k+1},Q^{k+1})$ is a post-processing of $(\Pbar{k},Q^k)$ (append an independent draw from $Q$ and uniformly permute), so $\privProfRem{\alloc{k}{P,Q}}(\eps)\in[0,1]$ is nonincreasing in $k$ by \cref{lem:hs-facts}. Truncating \cref{eq:main-profile} at any integer $m\ge1$ with $\prob{}{1+\Bin(n-1,p)<m}\le\beta$ and using $\xi\le(n-1)p\delta_0$ immediately gives \cref{eq:cutoff}. Since $1+\Bin(n-1,p)$ stochastically dominates $\Bin(n,p)$, the multiplicative Chernoff bound \citep[Section~2.2]{BLM13} yields this tail bound for $m=\max\{1,\lfloor pn-\sqrt{2pn\ln(1/\beta)}\rfloor\}$ (the event is empty when $m=1$). The add direction is identical. 
\end{proof}

If the only available assumption is that $\alg$ is $(\eps_0,\delta_0)$-DP, then $\privProfAdd{\alg}(\eps_0)\le\delta_0$, and \cref{thm:reduction} gives the simpler bound
\begin{equation*}
 \privProf{\shuf{n}{\alg}}(\eps)\le\privProf{\alloc{m}{P,Q}}(\eps)+2n\delta_0+\beta
 \qquad\text{for all }\eps\ge0 ,
\end{equation*}
with $m$ and $\beta$ as in \cref{eq:cutoff}. For $\eps\le\eps_0$ this follows from \cref{eq:cutoff} since $(1+e^{\eps})e^{-\eps_0}\le2$, while for $\eps\ge\eps_0$ it follows from the unamplified guarantee $\privProf{\shuf{n}{\alg}}(\eps)\le\delta_0$. Indeed, revealing the permutation reduces a pass to adaptive post-processing of a single invocation of $\alg$, so $\shuf{n}{\alg}$ is at least as private as $\alg$, and $(P,Q)$ dominates $\shuf{n}{\alg}$ in the remove direction and $(Q,P)$ in the add direction.

We note that the reduction is designed for the regime $\eps_0=O(1)$, in which the clone probability $p=e^{-\eps_0}$ is a constant. For pure DP algorithms with large $\eps_0$, \citet{FMT21} achieve a better dependence on $\eps_0$ by hiding the distinguished response among clones of both neighboring inputs rather than only of the null element.

\subsection{PLD accounting}\label{sec:pld-accounting}
\Cref{thm:reduction} bounds $\privProf{\shuf{n}{\alg}}(\eps)$ for a single shuffled pass, separately for each $\eps$. However, modern accurate privacy accounting for composition and subsampling requires the PLD of a dominating pair of distributions~\citep{DGKKM22,Google20,FS26pld}. Thus, to support such accounting (in particular, for multi-epoch DP-SGD), we construct a dataset-independent dominating pair and PLD for $\shuf{n}{\alg}$ from the PLDs of random allocation \citep{FS26pld,FS26msfft}. For simplicity, we leave discretization and tail truncation to standard PLD accounting methods \citep{DGKKM22,FS26pld}. Under the assumptions of \cref{thm:reduction}, let $(\Pmix,\Qmix)$ be the mixture pair in \cref{eq:mixture-pair} for $p=e^{-\eps_0}$, and let $\mu_k$ be the PLD of $(\Pbar{k},Q^k)$ (\cref{def:hs}). Because the revealed count $k$ has the same distribution $w_k$ under $\Pmix$ and $\Qmix$, the PLD of $(\Pmix,\Qmix)$ is simply $\mu=\sum_{k=1}^{n}w_k\mu_k$. When $\delta_0=0$ (as for pure DP algorithms), $(\Pmix,\Qmix)$ and $(\Qmix,\Pmix)$ already dominate $\shuf{n}{\alg}$ in the remove and add directions (\cref{lem:exact-mixture}), directly giving composable PLDs. When $\delta_0>0$, the shuffled outputs lie within total variation distance $\xi=1-(1-p\delta_0)^{n-1}$ of distributions dominated by $(\Pmix,\Qmix)$ (\cref{lem:exact-mixture,lem:extension}), so by the triangle inequality in \cref{lem:hs-facts}, their hockey-stick divergence at every $\kappa\ge0$ exceeds $\HockeyStickDiv{\kappa}{\Pmix}{\Qmix}$ by at most $(1+\kappa)\xi$. To construct a dominating pair and PLD in this case, we use the following general lemma, proved in \cref{app:reduction-proofs}.

\begin{lemma}[PLD under total-variation perturbation]\label{lem:pld-tv}
Given $\xi\in[0,1]$ and a pair of distributions $(P,Q)$ with PLD $\mu$, there exists a pair of distributions $(P^\xi,Q^\xi)$ with the following properties. For any pair $(P',Q')$ satisfying $\HockeyStickDiv{\kappa}{P'}{Q'}\le\HockeyStickDiv{\kappa}{P}{Q}+(1+\kappa)\xi$ for all $\kappa\ge0$, the pair $(P^\xi,Q^\xi)$ dominates $(P',Q')$. Moreover, $\HockeyStickDiv{\kappa}{P^\xi}{Q^\xi}\le\HockeyStickDiv{\kappa}{P}{Q}+(1+\kappa)\xi$ for every $\kappa\ge0$. Finally, the PLD of $(P^\xi,Q^\xi)$ depends only on $\mu$ and $\xi$, and can be computed from $\mu$ in $O(s)$ time when $\mu$ is supported on $s$ sorted points.
\end{lemma}

The pair $(P^\xi,Q^\xi)$ is obtained by trimming the upper tail of the privacy loss of $Q$-mass $\xi$ and its lower tail of $P$-mass $\xi$, and moving their $P$- and $Q$-masses into atoms of $+\infty$ and $-\infty$ privacy loss. Applying \cref{lem:pld-tv} to $(\Pmix,\Qmix)$ yields a dataset-independent pair $(P_p,Q_p)$ that dominates $\shuf{n}{\alg}$ in the remove direction; by \cref{lem:kov}, $(Q_p,P_p)$ dominates $\shuf{n}{\alg}$ in the add direction, and the PLDs of both pairs compose by convolution as above.

For a single-pass bound on $\privProf{\shuf{n}{\alg}}(\eps)$, one can optimize the clone probability $p=e^{-\eps_0}$ directly in \cref{eq:main-profile}, whereas we require a single dominating pair that tightly captures the privacy profile across all privacy-loss thresholds simultaneously. We achieve this approximately by folding the bounds for a finite set $\mathcal P\subset(0,1)$ of clone probabilities into a single PLD, using the lower convex hull of their hockey-stick curves \citep[Lemma~9]{ZDW22}. Specifically, denoting $\delta_0=\privProfAdd{\alg}(\ln(1/p))$ and $\xi_p\coloneqq1-(1-p\delta_0)^{n-1}$ for each $p\in\mathcal P$, we take the lower convex hull $H_*$ (\cref{sec:dom}) of
\[
 \kappa\mapsto\min\Bigl\{\HockeyStickDiv{\kappa}{P}{Q},\ \min_{p\in\mathcal P}\HockeyStickDiv{\kappa}{P_p}{Q_p}\Bigr\},\qquad \kappa\ge0,
\]
where $(P_p,Q_p)$ is the pair obtained above for clone probability $p$ with $\xi=\xi_p$, and the unamplified pair $(P,Q)$ dominates $\shuf{n}{\alg}$ by itself (\cref{sec:statement}). This yields a single \emph{envelope pair} $(P_*,Q_*)$ (note that the envelope is taken over hockey-stick curves, not PLD masses). By construction and \cref{lem:kov}, $(P_*,Q_*)$ dominates $\shuf{n}{\alg}$ in the remove direction and $(Q_*,P_*)$ dominates it in the add direction, giving single-pass PLDs (with $\xi_p$ already folded in) that compose directly across passes (\cref{app:implementation}). 

\section{Tight bounds for Gaussian shuffling}\label{sec:gauss-shuffle}
In this section we combine \cref{thm:reduction} with the bounds for Gaussian random allocation in \citep{FS26clt} to obtain asymptotically tight upper bounds for the shuffled Gaussian mechanism $\shuf{n}{\GaussMech{\sigma}}$ under zero-out adjacency. The full proof of \cref{thm:gauss-shuffle} appears in \cref{app:gauss-shuffle-proofs}. 

\begin{theorem}[Gaussian shuffling]\label{thm:gauss-shuffle}
Under zero-out adjacency, the shuffled Gaussian mechanism satisfies the following guarantees.
\begin{enumerate}
 \item There exist universal constants $C,C_1,C_2>0$ such that given $n\in\naturals$ with $n\ge2$, $\sigma>0$, and $\delta\in(0,1/2]$, if $\sigma\ge C_1\sqrt{\ln(n/\delta)}$ and $n\ge C_2\ln^2(n/\delta)$, then $\shuf{n}{\GaussMech{\sigma}}$ is $(\eps,\delta)$-DP for $\eps=C\sqrt{\ln(1/\delta)}/(\sigma\sqrt n)$.
 \item Given sequences $\sigma=\sigma(n)>0$ and $\delta=\delta(n)\in(0,1/2]$, if as $n\to\infty$ we have $\ln(1/\delta)=\omega(\ln(n))$, $\ln(1/\delta)=o(\sqrt n)$, and $\sigma=\omega\bigl(\sqrt{\ln(1/\delta)}\bigr)$, then $\shuf{n}{\GaussMech{\sigma}}$ is $(\eps,\delta)$-DP for $\eps=(1+o(1))\sqrt{2\ln(1/\delta)}/(\sigma\sqrt n)$.
\end{enumerate}
\end{theorem}
Both bounds match the random-allocation guarantees in~\citep[Corollaries~5.4 and~E.2]{FS26clt}: relative to~\citep[Corollary~5.4]{FS26clt}, the first claim adds the noise floor $\sigma\ge C_1\sqrt{\ln(n/\delta)}$ (with $n\ge C_2\ln^2(n/\delta)$ replacing the corollary's hypothesis), while relative to the second sufficient condition of~\citep[Corollary~E.2]{FS26clt}, the second claim strengthens $\sigma=\Omega\bigl(\sqrt{\ln(1/\delta)}\bigr)$ to $\sigma=\omega\bigl(\sqrt{\ln(1/\delta)}\bigr)$ (under which $e^{1/\sigma^2}-1=(1+o(1))/\sigma^2$). Like those guarantees, they match Poisson subsampling with sampling probability $1/n$ up to a constant factor in the first claim and up to a $1+o(1)$ factor in the second.

\begin{proof}[Proof outline]
By \cref{clm:gauss-dom} and the Gaussian tail bound on \cref{eq:gauss-profile}, choosing $\eps_0=O\bigl(\sqrt{\ln(n/\delta)}/\sigma\bigr)$ makes $\delta_0$ sufficiently small relative to $\delta/n$. The single-count bound \cref{eq:cutoff} then reduces both profiles of $\shuf{n}{\GaussMech{\sigma}}$ (up to a constant factor in $\delta$) to $1$-out-of-$m$ random allocation of the Gaussian pair, where $m$ is a high-probability lower bound on the clone count. In the first claim, $\sigma=\Omega(\sqrt{\ln(n/\delta)})$ gives $\eps_0=O(1)$ and hence a constant clone fraction ($m=\Omega(n)$ for sufficiently large $n$), at which point \citep[Corollary~5.4]{FS26clt} yields $\eps=O\bigl(\sqrt{\ln(1/\delta)}/(\sigma\sqrt n)\bigr)$. In the second claim, the faster noise growth forces $\eps_0\to0$ so almost every element acts as a clone ($m=(1-o(1))n$), and \citep[Corollary~E.2]{FS26clt} gives the $(1+o(1))$ bound.
\end{proof}

Both claims are asymptotically tight for sufficiently small $\delta$. Summing the $n$ responses implements the central-model Gaussian mechanism with noise scale $\sigma\sqrt n$ (for any view-independent $f_{\view}=f$ achieving the unit norm bound), which already requires $\eps=(1+o(1))\sqrt{2\ln(1/\delta)}/(\sigma\sqrt n)$ whenever $\ln(1/\delta)=\omega(\ln(\sigma\sqrt n))$. Under the hypotheses of the second claim this holds whenever $\ln(\sigma)=o(\ln(1/\delta))$; for example, for $\delta=e^{-n^a}$ with $0<a<1/2$, the second claim applies and is tight whenever $\sigma/n^{a/2}\to\infty$ and $\ln(\sigma)=o(n^a)$. Moreover, random allocation is itself the special case of shuffling in which all common elements are null, so any lower bound for random allocation (such as the numerical lower estimates in~\citep{FS26msfft}) is also a lower bound for shuffling, and can be tighter than the bound from the sum.

\section{Tighter lower bounds for Gaussian shuffling}\label{sec:lower-bound}
While our upper bounds for shuffling approach those of Poisson subsampling and random allocation at high noise, at low and moderate noise they are substantially higher. At low noise this is not merely a gap in the analysis: as depicted in \cref{fig:main}, our upper bound there is nearly matched by the lower bound in~\citep[Section~4.3]{CGKKMSZ24a}. That lower bound is based on a one-dimensional, view-independent Gaussian instance in which every common element contributes $-1$ and the distinguished element contributes $1$ when present and $0$ when absent (conjectured to be the worst case for shuffling; \citep[Conjecture~3.2]{CGKKMSZ24a}). We refer to this instance as the \emph{adversarial instance}. The prior lower bounds for this instance~\citep{CGKKMSZ24a,CGKKMSZ24b} use a test that thresholds the largest released response. This test is nearly optimal at very low noise but ignores the other $n-1$ responses, which carry most of the privacy loss at moderate and high noise. In this section we instead express the exact single-pass hockey-stick divergence of this instance through likelihood-ratio tests. For multiple passes, we compute the PLD of a pair of distributions dominated by a single pass. This PLD is a lower bound on the single-pass PLD that composes across passes. Proofs and computational details appear in \cref{app:lower-bound,app:implementation}.

Writing $P\coloneqq\Nrm(1,\sigma^2)$, $Q\coloneqq\Nrm(0,\sigma^2)$, and $G\coloneqq\Nrm(-1,\sigma^2)$, a single shuffled pass draws $n-1$ responses from $G$ and one response, at a uniformly random position, from $P$ on $(\dataset,x)$ and from $Q$ on $(\dataset,\bot)$. Because all common responses share the same distribution $G$, the output distributions $U_x$ and $U_\bot$ of $z=(z_1,\ldots,z_n)$ collapse to uniform mixtures over the position of the distinguished response, with likelihood ratio
\begin{equation}\label{eq:identical-ratio}
 \frac{U_x(z)}{U_\bot(z)}=\frac{\sum_{j=1}^nP(z_j)/G(z_j)}{\sum_{j=1}^nQ(z_j)/G(z_j)}.
\end{equation}
This likelihood-ratio identity (\cref{eq:identical-ratio}) already explains the low-noise gap with random allocation: writing $P(z_j)/G(z_j)=\bigl(P(z_j)/Q(z_j)\bigr)\bigl(Q(z_j)/G(z_j)\bigr)$ expresses the ratio as an average of the single-step ratios $P(z_j)/Q(z_j)$, weighted by the posterior probability under $U_\bot$ that $z_j$ is the distinguished response (proportional to $Q(z_j)/G(z_j)$). In random allocation ($G=Q$) these weights are uniform, whereas at low noise they concentrate on responses much more likely under $Q$ than under $G$. Moreover, \cref{eq:identical-ratio} reduces the optimal single-pass likelihood-ratio test to thresholding a sum of $n$ independent scalar terms, as formalized in the following lemma (specializing \citep[Theorem~14]{SCW25}; proved in \cref{app:lower-bound}).

\begin{lemma}[Single-pass likelihood-ratio test]\label{lem:identical-ratio}
Given $n\in\naturals$, $\sigma>0$, and $\kappa\ge0$, the shuffled output distributions $U_x$ and $U_\bot$ of the adversarial Gaussian instance satisfy
\begin{equation}\label{eq:identical-hs}
 \HockeyStickDiv{\kappa}{U_x}{U_\bot}
 =\prob{z\sim U_x}{\sum_{j=1}^n\Delta_\kappa(z_j)>0}-\kappa\,\prob{z\sim U_\bot}{\sum_{j=1}^n\Delta_\kappa(z_j)>0},
\end{equation}
where $P\coloneqq\Nrm(1,\sigma^2)$, $Q\coloneqq\Nrm(0,\sigma^2)$, $G\coloneqq\Nrm(-1,\sigma^2)$, $U_x\coloneqq\frac1n\sum_{j=1}^n G^{j-1}\times P\times G^{n-j}$, $U_\bot\coloneqq\frac1n\sum_{j=1}^n G^{j-1}\times Q\times G^{n-j}$, and $\Delta_\kappa\coloneqq(P-\kappa Q)/G$. Moreover, under $z\sim U_x$ the sum $\sum_{j=1}^n\Delta_\kappa(z_j)$ has the same distribution as $\sum_{j=1}^{n-1}\Delta_\kappa(y_j)+\Delta_\kappa(y_n)$ for independent $y_1,\ldots,y_{n-1}\sim G$ and $y_n\sim P$, and under $z\sim U_\bot$ the same holds with $y_n\sim Q$.
\end{lemma}

By \cref{lem:identical-ratio}, computing the exact single-pass hockey-stick divergence in the remove direction is a convolution task over $n$ independent scalar terms (the add direction is analogous, with $(Q-\kappa P)/G$ in place of $\Delta_\kappa$ and the roles of $U_x$ and $U_\bot$ exchanged).

Over $E$ shuffled passes with independent permutations, the output distributions are $U_x^E$ and $U_\bot^E$, and the optimal test no longer thresholds a single sum of independent terms. Instead, we obtain lower bounds on the $E$-pass hockey-stick divergence by composing a discrete pair of distributions dominated by a single pass. We construct this pair from lower bounds on the probabilities of finitely many single-pass test events under $U_x$ and upper bounds on their probabilities under $U_\bot$, using the following lemma (based on \citep[Lemma~17]{BR14}; see also \citep[Lemma~9]{ZDW22} and \citep[Lemma~3.2]{DGKKM22}; proved in \cref{app:lower-bound}).

\begin{lemma}[Dominated pair from test bounds]\label{lem:pool}
Given $r\in\naturals$, $p_1,\ldots,p_r,q_1,\ldots,q_r\in[0,1]$, distributions $R,S$ over a common space, and events $A_1,\ldots,A_r$, if $p_i\le R(A_i)$ and $q_i\ge S(A_i)$ for every $i\in[r]$ with $p_i=0$ whenever $q_i=0$, then there exists a discrete pair of distributions $(\widetilde P,\widetilde Q)$ over $[s]$ (for some $s\le r+1$) with $\widetilde Q(j)>0$ for all $j\in[s]$ that is dominated by $(R,S)$ and satisfies
\[
 \HockeyStickDiv{\kappa}{\widetilde P}{\widetilde Q}
 =\max\bigl(\{0,1-\kappa\}\cup\{p_i-\kappa q_i:i\in[r]\}\bigr)
 \qquad\text{for all }\kappa\ge0.
\]
Moreover, $(\widetilde P,\widetilde Q)$ depends only on $\{(q_i,p_i):i\in[r]\}$ and is given by the increments between consecutive vertices of the upper concave hull of $\{(0,0),(1,1)\}\cup\{(q_i,p_i):i\in[r]\}$.
\end{lemma}

Applying \cref{lem:pool} to $(R,S)=(U_x,U_\bot)$ with a finite family of single-pass test events yields a discrete pair $(\widetilde P,\widetilde Q)$ dominated by $(U_x,U_\bot)$. By composition of dominating pairs (\cref{sec:dom}), $(U_x^E,U_\bot^E)$ then dominates $(\widetilde P^E,\widetilde Q^E)$, whose PLD is the $E$-fold convolution of the PLD of $(\widetilde P,\widetilde Q)$ and therefore lower-bounds the $E$-pass hockey-stick divergence of $\shuf{n}{\GaussMech{\sigma}}$ (\cref{eq:multipass-lb} in \cref{app:lower-bound}). To compute rigorous one-sided test bounds across all thresholds at low noise, we shift $\Delta_\kappa$ by a constant that makes every summand nonnegative via completing the square on the Gaussian densities, enabling multiscale FFT convolution \citep{FS26msfft} (\cref{app:lower-bound}). Because these shifted tests include the exact likelihood-ratio events of \cref{lem:identical-ratio}, taking them over a dense set of thresholds $\kappa$ approaches the exact $E$-pass divergence of this instance (up to discretization).

\section{Numerical evaluation}\label{sec:evaluation}
In this section we evaluate the privacy accountant for shuffling given by \cref{thm:reduction,sec:pld-accounting} across three settings: the Gaussian mechanism (\cref{sec:numerics}), pure DP algorithms (\cref{sec:pure-clones}), and $20$ published multi-epoch DP-SGD runs (\cref{sec:dpsgd}). For the Gaussian mechanism and the DP-SGD runs, we also compare it with our lower bounds (\cref{sec:lower-bound}). Given a pair $(P,Q)$ that dominates $\alg$ and a clone probability $p=e^{-\eps_0}\in(0,1)$ with $\delta_0=\privProfAdd{\alg}(\eps_0)$, we compute the allocation profiles and PLDs from $(P,Q)$ using the algorithms in~\citep{FS26pld,FS26msfft}. For the Gaussian mechanism we combine the single-pass dominating pairs across a grid of clone probabilities into a single envelope pair (\cref{sec:pld-accounting}). We query its PLD directly for a single shuffled pass and compose it across epochs for multi-epoch runs. For the pure DP algorithms we use the largest clone probability with $\delta_0=0$ and evaluate the count-mixture bound \cref{eq:main-profile} in both directions.

For replacement adjacency we take the smaller of the group bound (\cref{lem:group}) and the neighbor-reference bound (\cref{lem:neighbor-reference}) derived in \cref{app:replacement}. We describe the implementation details in \cref{app:experiments}; the code of all experiments in this work is available online~\citep{FS26code}.

\subsection{The Gaussian mechanism}\label{sec:numerics}
\begin{figure}[t]
 \centering
 \includegraphics[width=\linewidth]{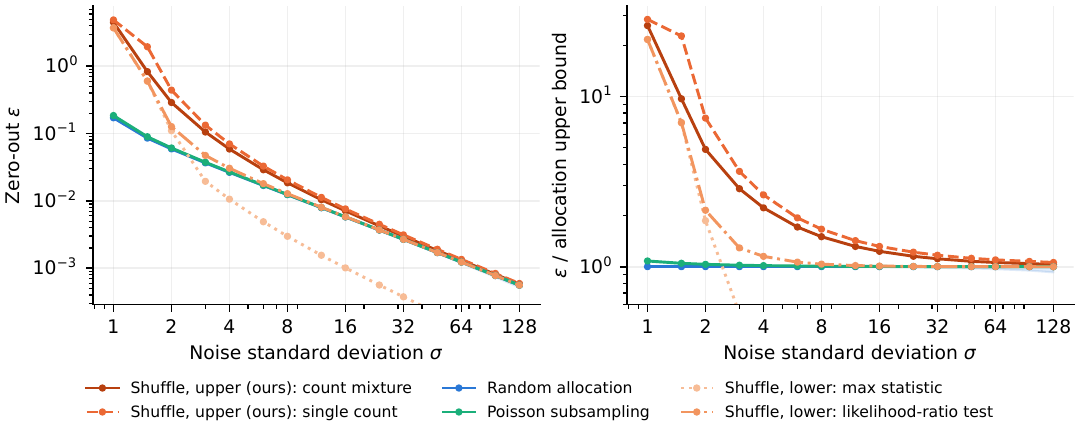}
 \caption{$\eps$ of one shuffled pass of the Gaussian mechanism with $n=1000$, $\delta=10^{-6}$, unit norm bound, and zero-out adjacency, in absolute terms (left) and relative to the random allocation upper bound (right). Our upper bounds are the PLD envelopes obtained from \cref{eq:main-profile} (count mixture) and \cref{eq:cutoff} (single count), and the lower bounds are those in~\citep{CGKKMSZ24a} (max statistic) and \cref{lem:identical-ratio} (likelihood-ratio test). }
 \label{fig:comparison}
\end{figure}

We compare one shuffled pass $\shuf{n}{\GaussMech{\sigma}}$ of the Gaussian mechanism with $1$-out-of-$t$ random allocation $\alloc{t}{P,Q}$ and $t$-step Poisson subsampling of the Gaussian pair $(P,Q)$, at $t=n=1000$, $\delta=10^{-6}$, and unit norm bound, all under zero-out adjacency (\cref{fig:comparison}). The Poisson scheme has sampling probability $1/t$, so that in all three every element is used once in expectation. For shuffling we apply the accountant with the Gaussian pair (\cref{clm:gauss-dom}), for which $\privProfAdd{\GaussMech{\sigma}}(\eps_0)$ is given by \cref{eq:gauss-profile}. We report the PLD envelopes obtained from the count-mixture bound \cref{eq:main-profile} and the single-count bound \cref{eq:cutoff}, each combining $1001$ clone probabilities (\cref{app:implementation}). The shaded numerical brackets for random allocation and Poisson subsampling in \cref{fig:comparison} quantify discretization error rather than looseness of the shuffling reduction.

At high noise almost every response can serve as a clone, and the shuffling bound approaches random allocation. It is within about $\resAllocGapHigh\%$ of the allocation bound for $\sigma\ge32$ and within a few percent at $\sigma=128$, where clone probabilities above $\resCloneProbHigh$ are available. Random allocation and Poisson subsampling differ by less than $\resPoissonGap\%$, and by less than $\resPoissonGapHigh\%$ for $\sigma\ge12$, matching \citep{FS26pld}. At moderate noise the gap grows, to a factor of about $\resAllocRatioEight$ at $\sigma=8$ and $\resAllocRatioFour$ at $\sigma=4$, and at $\sigma\le2$ the shuffling bound exceeds the allocation bound by a factor of about $\resAllocRatioLow$ or more. Keeping the full count mixture instead of the single-count bound matters most at low noise, where few clones are available; for example, at $\sigma=1.5$ the mixture bound is less than half of the single-count bound.

\para{Zero-out lower bounds.}
\Cref{fig:comparison} also compares our upper bounds with two lower bounds on the adversarial instance of \cref{sec:lower-bound}. The dotted curve is the max-statistic bound in~\citep{CGKKMSZ24a}, which already gives $\eps\ge\resMaxStatSigmaOne$ at $\sigma=1$ against our upper bound of $\resUpperSigmaOne$, but loses tightness at moderate noise. The dash-dotted curve is our single-pass likelihood-ratio lower bound (\cref{lem:identical-ratio}), computed via FFT convolution (\cref{app:lower-bound,app:implementation}). It nearly coincides with the max-statistic bound for $\sigma\le1.5$, improves on it by about $\resLRGainTwo\%$ at $\sigma=2$ and by more than a factor of $\resLRGainEight$ at $\sigma=8$, and strictly exceeds the upper bound for random allocation for all $\sigma\le16$ (only slightly for $\sigma\ge12$). For $\sigma\ge24$ it matches that bound within a fraction of a percent. Our upper bound stays within a factor of $\resUpperLRMax$ of this lower bound at every noise level, and within $\resUpperLRLow$ for $\sigma\le1.5$. Thus, much of the low-noise gap between shuffling and random allocation is inherent to shuffling, as explained by the nonuniform posterior weighting in the likelihood-ratio identity (\cref{eq:identical-ratio}). 

\begin{figure}[t]
\centering
\includegraphics[width=\linewidth]{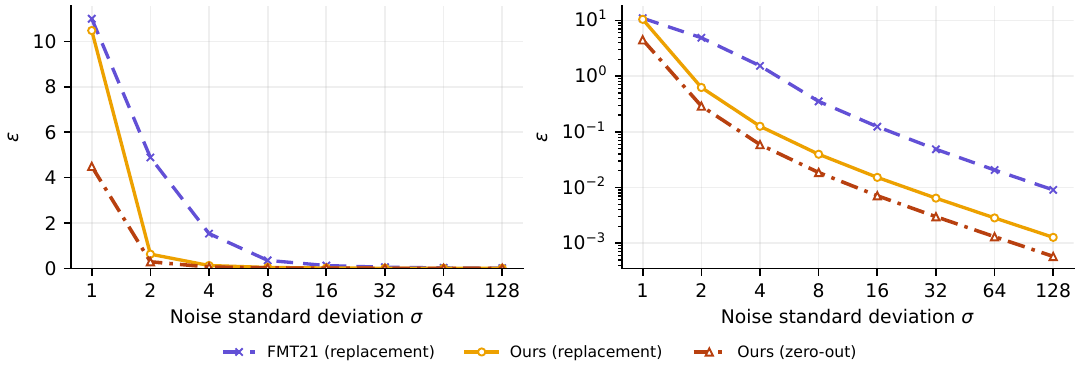}
\caption{Replacement bounds for one shuffled pass of the Gaussian mechanism with $n=1000$, $\delta=10^{-6}$, and unit norm bound, on a linear (left) and logarithmic (right) scale. The baseline is the clones bound in~\citep[Theorem~3.8]{FMT21} (capped at the unamplified guarantee), our replacement bound is the smaller of the group and neighbor-reference bounds, and our zero-out bound is shown for reference.}
\label{fig:gauss-replacement}
\end{figure}

\para{Replacement adjacency.}
Under replacement adjacency, the response means of two neighboring inputs can differ by up to $2$, so the pair $(\Nrm(2,\sigma^2),\Nrm(0,\sigma^2))$ dominates $\GaussMech{\sigma}$ \citep[Lemma~3]{BW18}. By scaling, the neighbor-reference bound (\cref{lem:neighbor-reference}) behaves like the zero-out count mixture at half the noise scale, $\sigma/2$, which sharply reduces the clone probability at low and moderate noise. The group bound (\cref{lem:group}) instead applies the zero-out bound twice at the full noise scale $\sigma$, keeping many more clones at only roughly a factor-of-$2$ cost in $\eps$. Consequently, the group bound is lower at every noise level (by a factor of about $\resGroupNRTwo$ at $\sigma=2$, decreasing as the noise grows; \cref{fig:replacement-comparison} in \cref{app:replacement}) and remains slightly above twice the zero-out bound. As shown in \cref{fig:gauss-replacement}, our replacement bound improves on the clones bound in~\citep[Theorem~3.8]{FMT21} (applied to an $(\eps_0,\delta_0)$-DP guarantee of the algorithm) by a factor of $\resClonesGain$ or more for $2\le\sigma\le128$, whereas at $\sigma=1$ all replacement bounds are close to the unamplified guarantee. 

\subsection{Pure DP algorithms}\label{sec:pure-clones}
\begin{figure}[!htbp]
\centering
\includegraphics[width=\linewidth]{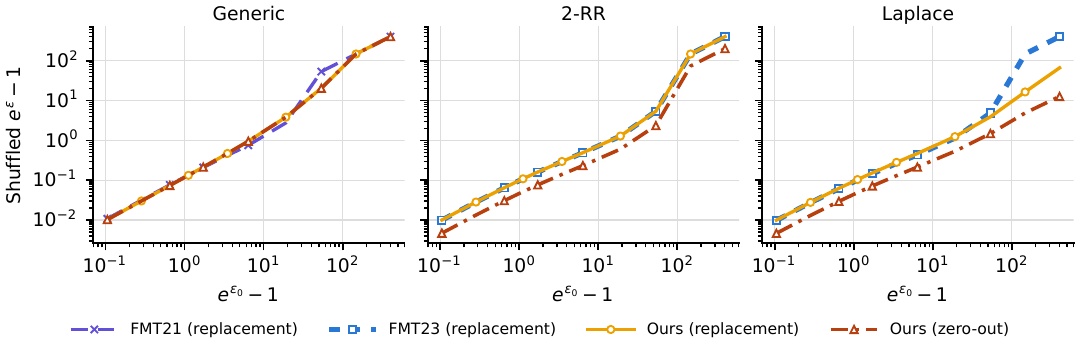}
\caption{Shuffled $e^{\eps}-1$ against $e^{\eps_0}-1$ for pure DP algorithms, with $n=1000$ and $\delta=10^{-6}$. For the generic algorithm our zero-out and replacement bounds coincide. The clones bound in~\citep{FMT21} depends only on $\eps_0$ and is shown only in the generic panel.}
\label{fig:pure-clones}
\end{figure}

Although our reduction is motivated primarily by the Gaussian mechanism, it also applies directly to pure $\eps_0$-DP algorithms (with zero filling error, $\delta_0=0$). We compare our bounds with the generic pure DP shuffling bound in~\citep{FMT21} and the algorithm-specific bounds in~\citep{FMT23} on three benchmarks: a generic $\eps_0$-DP algorithm, binary randomized response, and the Laplace mechanism (\cref{fig:pure-clones}, with $n=1000$ and $\delta=10^{-6}$). Because the bounds in~\citep{FMT21,FMT23} hold for replacement adjacency (which implies the zero-out guarantee in all three examples), we compare them both with our zero-out bounds (favoring our bounds, since replacement is the stronger guarantee) and directly with our replacement bounds. We give the computation and decomposition details in \cref{app:pure-settings}.

\para{Benchmarks.}
A generic pure $\eps_0$-DP algorithm satisfies $\alg(x,\view)\ge e^{-\eps_0}\alg(x',\view)$ for all $x,x'$, and every pair of responses is dominated by the binary randomized-response pair $P=(e^{\eps_0},1)/(e^{\eps_0}+1)$ and $Q=(1,e^{\eps_0})/(e^{\eps_0}+1)$. Without algorithm-specific structure, $\bot$ responds as a fixed element of $\domain$, so every zero-out comparison is a replacement comparison, and the neighbor-reference bound (\cref{lem:neighbor-reference}, at clone probability $p=e^{-\eps_0}$) gives both our zero-out and replacement bounds (the group bound cannot be lower here; \cref{app:replacement}).

Binary randomized response outputs its input $a\in\{0,1\}$ with probability $e^{\eps_0}/(e^{\eps_0}+1)$ and $1-a$ otherwise (distribution $P_a$). Its neighbor-reference bound coincides with the generic bound at clone probability $e^{-\eps_0}$. Under zero-out adjacency, however, taking the uniform distribution $U=\frac12(P_0+P_1)$ as the null response places $\bot$ midway between $0$ and $1$: with dominating pair $(P_a,U)$, each $P_a$ contains a $2/(e^{\eps_0}+1)$-fraction of $U$, increasing the clone probability from $e^{-\eps_0}$ to $2/(e^{\eps_0}+1)$.

For the Laplace mechanism $\alg(x)=x+\operatorname{Lap}(1/\eps_0)$ on $x\in[0,1]$ (where $\operatorname{Lap}(1/\eps_0)$ is the Laplace distribution with scale $1/\eps_0$), any two inputs are separated by at most $1$, so the neighbor-reference bound uses the unit-displacement pair $(\alg(1),\alg(0))$ at clone probability $e^{-\eps_0}$. Under zero-out adjacency, placing the null response at the midpoint $\alg(1/2)$ halves the worst-case displacement to $1/2$ and increases the clone probability from $e^{-\eps_0}$ to $e^{-\eps_0/2}$ (\cref{app:pure-settings}).

\para{Comparison.}
In the zero-out comparison, our bounds for randomized response and the Laplace mechanism improve on those in~\citep{FMT23} at every $\eps_0$ we evaluate, reflecting the larger clone probabilities enabled by their midpoint null responses. The bound in~\citep{FMT21} does not improve under zero-out adjacency, since it depends on the algorithm only through $\eps_0$; at $\eps_0=2$, our zero-out bounds for randomized response and the Laplace mechanism are lower than it by roughly a factor of $\resPureFMTGainTwo$.

In the replacement comparison (where all percentages are relative differences of $\eps$), the group bound is lower than the neighbor-reference bound for randomized response and the Laplace mechanism at every $\eps_0\le4$ (\cref{app:replacement}), and lower than the bound in~\citep{FMT21} throughout this range. Relative to the algorithm-specific bounds in~\citep{FMT23}, it stays within $\resRRFMTWithin\%$ for binary randomized response, and for the Laplace mechanism it is at most about $\resLaplaceFMTAbove\%$ higher for $\eps_0\le3$ and lower for $\eps_0\ge4$. For the generic algorithm, where the group bound cannot help, our bound is lower than that in~\citep{FMT21} for $\eps_0\le\resGenericBelowMax$ and at $\eps_0=4$ (where the bound in~\citep{FMT21} nears $\eps_0$), and higher by up to about $\resGenericAbovePct\%$ for $\resGenericAboveMin\le\eps_0\le\resGenericAboveMax$. This reflects the better dependence on $\eps_0$ of the analysis in~\citep{FMT21}, which uses clones of both neighboring inputs (\cref{sec:statement}).

\subsection{Published DP-SGD settings}\label{sec:dpsgd}
We now evaluate the tightness of our bound in settings used in practice, taking $20$ DP-SGD training runs from seven well-cited papers \citep{BPS19,BDLS20,TB21,KSCGTT22,YNBGIKKLMWYZ22,BDSHSSKSB23,LZLRMS23}. All seven papers report privacy guarantees for Poisson subsampling, whereas in $\resShuffledSettings$ of the $20$ settings the released code trains with shuffled or fixed batches. The settings span models from logistic regression to pretrained language models across image, text, tabular, and recommendation datasets of up to $\resMaxExamplesMillions$ million examples, with reported $\eps$ between $\resMinReportedEps$ and $\resMaxReportedEps$. From each run we take the dataset size $N$, batch size $B$, number of steps $T$, and reported $(\eps,\delta)$ (\cref{tab:dpsgd-settings} in \cref{app:dpsgd}), and model it as a shuffled run of $E=\lceil T/n\rceil$ full epochs over $n=\lfloor N/B\rfloor$ batches of size $B$. For each setting we compute the noise multiplier $\sigma$ (noise standard deviation divided by clipping norm) that our upper bound shows to be sufficient for the reported $(\eps,\delta)$, together with the $\sigma$ shown to be necessary by the max-statistic lower bound \citep{CGKKMSZ24b} and by our multi-pass lower bound (\cref{sec:lower-bound,app:lower-bound}). All three bounds extend to batches and multiple epochs with independent permutations (\cref{sec:pld-accounting,app:dpsgd}). As a baseline, we also compute the \emph{matching} Poisson $\sigma$, namely the noise multiplier required by Poisson subsampling with $En$ steps and sampling probability $1/n$.

\begin{figure}[t]
\centering
\includegraphics[width=\linewidth]{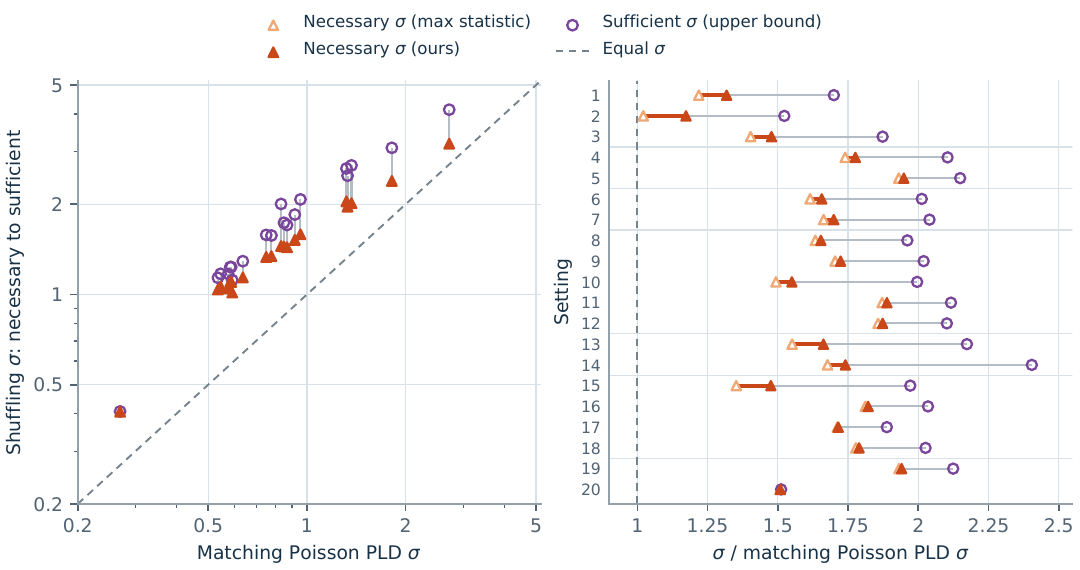}
\caption{Noise multiplier $\sigma$ required for the reported $(\eps,\delta)$ in the settings of \cref{tab:dpsgd-settings}, for the max-statistic lower bound \citep{CGKKMSZ24b}, our lower bound (\cref{sec:lower-bound}), and our upper bound. Left: our necessary and sufficient $\sigma$ for shuffling against the matching Poisson $\sigma$, on logarithmic axes. Right: the necessary and sufficient $\sigma$ divided by the matching Poisson $\sigma$, for each setting.}
\label{fig:dpsgd}
\end{figure}

\Cref{fig:dpsgd} plots the necessary and sufficient $\sigma$ for each setting against the matching Poisson $\sigma$ (left) and relative to it (right), with exact values in \cref{tab:dpsgd-noise} (\cref{app:dpsgd}). Across all $20$ settings, our upper bound closely tracks our lower bound. The sufficient $\sigma$ is at most $\resSufNecMax$ times the necessary $\sigma$, with a median ratio of $\resSufNecMedian$. The ratio is smallest at larger $\eps$ (at most $\resSufNecMaxLargeEps$ whenever $\eps\ge4.6$) and rises modestly at smaller $\eps$, where $\sigma$ lies in the moderate-noise regime of \cref{fig:comparison}. In setting $\resUnampSetting$ (reported $\eps=\resMaxReportedEps$), both bounds match the unamplified $\sigma$ to within $\resUnampPct\%$, and thus shuffling gives essentially no amplification there. Combining multiple clone probabilities (\cref{sec:pld-accounting}) lowers the sufficient $\sigma$ by a median of $\resEnvelopeMedian\%$ and up to about $\resEnvelopeMaxAbout\%$ compared with using a single clone probability (\cref{app:dpsgd}).

Compared with the max-statistic bound \citep{CGKKMSZ24b}, our lower-bound accountant (\cref{app:lower-bound}) increases the necessary $\sigma$ in $\resNecGainCount$ of the $20$ settings (and matches it in setting $\resNecNoGainSetting$), with the largest gains at moderate noise (e.g., from $\resNecGainFrom$ to $\resNecGainTo$ in setting $\resNecGainSetting$). The lower bound also shows that in most of these settings shuffling is substantially less private than Poisson subsampling. In $\resNecMatchCount$ of the $20$ settings our necessary $\sigma$ for shuffling is between $\resNecMatchMin$ and $\resNecMatchMax$ times the matching Poisson $\sigma$. In the other two settings (settings $\resNecMatchLowFirstID$ and $\resNecMatchLowSecondID$) it is $\resNecMatchLowFirst$ and $\resNecMatchLowSecond$ times the matching Poisson $\sigma$, respectively. These are the settings with the largest matching Poisson $\sigma$, in agreement with the smaller gap at high noise in \cref{sec:numerics}. At the same time, our upper bound shows that $\resSufMatchMin$ to $\resSufMatchMax$ times the matching Poisson $\sigma$ (at most $\resSufMatchSecond$ in $19$ of the $20$ settings) suffices for the reported $(\eps,\delta)$ under shuffling.

The seven papers use four different accountants for their reported Poisson guarantees: R\'enyi DP, PLD, the privacy random variable (PRV) accountant \citep{GLW21}, and the central limit approximation of Gaussian DP \citep{DRS22}. We recomputed $\sigma$ for Poisson subsampling with a PLD accountant \citep{Google20}, using the sampling probability and the number of steps reported in each paper (the last two columns of \cref{tab:dpsgd-settings}). Compared with the reported $\sigma$, the recomputed $\sigma$ is lower in the settings accounted with R\'enyi DP or the PRV accountant, higher in the settings that use the central limit approximation (which does not give an upper bound on $\eps$), and about the same in the two settings accounted with PLD. The reported sampling probabilities and numbers of steps do not always correspond to a shuffled run with full batches, and the matching Poisson $\sigma$ differs from the recomputed $\sigma$ by at most $\resMatchChange\%$ (\cref{app:dpsgd}).

\section*{AI Usage}
The authors formulated the question and developed a solution using a different reduction to random allocation. The key ideas for the simplified and more efficient reduction presented here were found by ChatGPT Astra 6. Further development of these ideas and writing were performed through an iterative process in which frontier AI models generated details of proofs and wrote parts of the prose based on author-provided guidance and structure. The resulting text was also extensively edited and proofread by the authors. The implementation of the evaluation code was written by AI under close supervision and auditing by the authors. The authors take full responsibility for the content of this work.
We also note that we are aware of a recent AI-generated manuscript on privacy accounting for shuffling that is posted online. It appears to contain related ideas and is independent of our work. Given the quality of presentation, we do not comment further on its contents.
\clearpage

\appendix
\crefalias{section}{appendix}
\crefalias{subsection}{appendix}

\section{\texorpdfstring{Proofs for \cref{sec:preliminaries,sec:reduction}}{Proofs for Sections 2 and 3}}\label{app:reduction-proofs}
In this appendix we prove \cref{lem:hs-facts}, the single-step filling bound (\cref{lem:floor}) used in \cref{lem:extension}, and \cref{lem:pld-tv} on correcting a PLD for a total-variation perturbation.

\begin{proofof}{\cref{lem:hs-facts}}
The first property is the data-processing inequality for $f$-divergences \citep{SV16}. For the second, the variational form $\HockeyStickDiv{\kappa}{P}{Q}=\sup_{A}(P(A)-\kappa Q(A))$ gives $\HockeyStickDiv{\kappa}{P}{Q}\in[0,1]$ (via $A=\emptyset$ and $P(A)-\kappa Q(A)\le P(A)\le1$) and shows that $\kappa\mapsto\HockeyStickDiv{\kappa}{P}{Q}$ is convex and nonincreasing as the pointwise supremum of affine nonincreasing functions. To prove the third we use the same variational form together with $P(A)\le P'(A)+\TV(P,P')$ and $Q(A)\ge Q'(A)-\TV(Q,Q')$ for every event $A$. The fourth holds since the two mixtures have disjoint supports over the index $i$.
\end{proofof}

The next lemma formalizes the single-step filling construction used in the proof of \cref{lem:extension}.

\begin{lemma}[Filling the reference component]\label{lem:floor}
Given $p\in(0,1)$ and distributions $P,Q$ over $\outDom$,
\[
 r(\omega)\coloneqq\frac{[P(\omega)-pQ(\omega)]_+}{1-p+b}\qquad\text{and}\qquad
 \widetilde P\coloneqq pQ+(1-p)r
\]
define probability distributions over $\outDom$ such that $\widetilde P$ contains a $p$-fraction of $Q$ and $\TV(P,\widetilde P)=b$, where $b\coloneqq\int[pQ(\omega)-P(\omega)]_+\,d\omega$. Moreover, if $p=e^{-\eps_0}$ and $\HockeyStickDiv{e^{\eps_0}}{Q}{P}\le\delta_0$ for $\eps_0>0$ and $\delta_0\in[0,1]$, then $b\le p\delta_0$.
\end{lemma}
\begin{proof}
The function $P-pQ$ integrates to $1-p$ and its negative part integrates to $b$, so its positive part integrates to $1-p+b$. Thus $r$ is a probability distribution, and so is $\widetilde P$, which contains a $p$-fraction of $Q$. To compute the total variation distance we observe that on the event $\{P<pQ\}$ we have $\widetilde P=pQ>P$, while on its complement
\[
 \widetilde P=pQ+\frac{1-p}{1-p+b}\,(P-pQ)\le pQ+(P-pQ)=P .
\]
Thus the mass that $\widetilde P$ adds relative to $P$ is exactly $\int[pQ-P]_+=b$, so $\TV(P,\widetilde P)=b$. Finally, when $p=e^{-\eps_0}$, \cref{def:hs} gives $b=p\,\HockeyStickDiv{1/p}{Q}{P}=p\,\HockeyStickDiv{e^{\eps_0}}{Q}{P}\le p\delta_0$.
\end{proof}

\begin{proofof}{\cref{lem:pld-tv} (PLD under total-variation perturbation)}
For $\xi=0$ we take $(P^0,Q^0)\coloneqq(P,Q)$, whose PLD is $\mu$. For $\xi\in(0,1]$, let $H$ be an upper level set of $\lossFunc{\cdot}{P}{Q}$ with $Q(H)=\xi$ and let $D$ be a lower level set with $P(D)=\xi$, where we allow splitting a boundary level by adjoining an independent uniform random variable on $[0,1]$ to both distributions. If $P(H)+\xi\ge1$, we let $(P^\xi,Q^\xi)$ be any pair of distributions with disjoint supports, whose PLD $\mu_\xi=\operatorname{Dirac}_{+\infty}$ is a point mass at $+\infty$. Otherwise $P(H)+P(D)=P(H)+\xi<1$, and thus $H$ and $D$ can be chosen disjoint. Their likelihood-ratio thresholds $\kappa_H$ and $\kappa_D$ satisfy $0<\kappa_D\le\kappa_H<\infty$, with $P\ge\kappa_HQ$ on $H$, $P\le\kappa_DQ$ on $D$, and $\kappa_DQ\le P\le\kappa_HQ$ outside $H\cup D$. Writing $C\coloneqq(H\cup D)^c$, we define
\begin{equation}\label{eq:shuffle-corrected-pair}
 P^\xi\coloneqq P|_C+(P(H)+\xi)\operatorname{Dirac}_{+},\qquad
 Q^\xi\coloneqq Q|_C+(Q(D)+\xi)\operatorname{Dirac}_{-},
\end{equation}
where $P|_C$ and $Q|_C$ are unnormalized restrictions and $\operatorname{Dirac}_{+},\operatorname{Dirac}_{-}$ are point masses at two distinct new outcomes. Since $P(D)=Q(H)=\xi$, both distributions have total mass $1$, and the PLD of $(P^\xi,Q^\xi)$ is $\mu_\xi=\mu_C+(P(H)+\xi)\operatorname{Dirac}_{+\infty}$, where $\mu_C$ is the image of $P|_C$ under $\lossFunc{\cdot}{P}{Q}$. Because the distribution of $\lossFunc{\cdot}{P}{Q}$ under $Q$ has finite part $e^{-\loss}\mu(d\loss)$ and its remaining mass at $-\infty$, $\mu_\xi$ depends only on $\mu$ and $\xi$: when $\mu$ is supported on $s$ sorted points, scanning the grid and trimming the upper tail of $e^{-\loss}\mu(\loss)$-mass $\xi$ and lower tail of $\mu$-mass $\xi$ computes $\mu_\xi$ in $O(s)$ time.

Now suppose that $\HockeyStickDiv{\kappa}{P'}{Q'}\le\HockeyStickDiv{\kappa}{P}{Q}+(1+\kappa)\xi$ for every $\kappa\ge0$. When $\xi=0$ or $P(H)+\xi\ge1$ the claim is immediate, since in the latter case $\HockeyStickDiv{\kappa}{P^\xi}{Q^\xi}=1\ge\HockeyStickDiv{\kappa}{P'}{Q'}$ while $\HockeyStickDiv{\kappa}{P}{Q}\ge P(H)-\kappa Q(H)=P(H)-\kappa\xi\ge1-(1+\kappa)\xi$. Suppose therefore that $\xi>0$ and $P(H)+\xi<1$. For $\kappa\in[\kappa_D,\kappa_H]$, the upper level set $\{P>\kappa Q\}$ contains $H$ (up to boundary tie-breaking) and is disjoint from $D$, so by \cref{eq:shuffle-corrected-pair},
\begin{align*}
 \HockeyStickDiv{\kappa}{P^\xi}{Q^\xi}
 &=(P(H)+\xi)+\bigl(\HockeyStickDiv{\kappa}{P}{Q}-P(H)+\kappa\xi\bigr)\\
 &=\HockeyStickDiv{\kappa}{P}{Q}+(1+\kappa)\xi
 \ge\HockeyStickDiv{\kappa}{P'}{Q'}.
\end{align*}
For $\kappa>\kappa_H$, no point of $C\cup\{-\}$ has ratio above $\kappa_H$, so monotonicity of hockey-stick divergence in $\kappa$ (\cref{lem:hs-facts}) gives
\[
 \HockeyStickDiv{\kappa}{P'}{Q'}
 \le\HockeyStickDiv{\kappa_H}{P'}{Q'}
 \le\HockeyStickDiv{\kappa_H}{P^\xi}{Q^\xi}
 =P(H)+\xi
 =\HockeyStickDiv{\kappa}{P^\xi}{Q^\xi},
\]
and $P(H)+\xi=(P(H)-\kappa\xi)+(1+\kappa)\xi\le\HockeyStickDiv{\kappa}{P}{Q}+(1+\kappa)\xi$. Finally, for $\kappa\in[0,\kappa_D]$, every point of $\{+\}\cup C$ has ratio at least $\kappa_D\ge\kappa$, so $\HockeyStickDiv{\kappa}{P^\xi}{Q^\xi}=1-\kappa Q(C)$ is affine in $\kappa$. Because $\HockeyStickDiv{\kappa}{P'}{Q'}$ is convex in $\kappa$ (\cref{lem:hs-facts}) and bounded by $1-\kappa Q(C)$ at both $\kappa=0$ and $\kappa=\kappa_D$, we obtain $\HockeyStickDiv{\kappa}{P'}{Q'}\le\HockeyStickDiv{\kappa}{P^\xi}{Q^\xi}$ on $[0,\kappa_D]$ as well, while $1-\kappa Q(C)=(P(D^c)-\kappa Q(D^c))+(1+\kappa)\xi\le\HockeyStickDiv{\kappa}{P}{Q}+(1+\kappa)\xi$. Thus $(P^\xi,Q^\xi)$ dominates $(P',Q')$ and satisfies $\HockeyStickDiv{\kappa}{P^\xi}{Q^\xi}\le\HockeyStickDiv{\kappa}{P}{Q}+(1+\kappa)\xi$ for all $\kappa\ge0$.
\end{proofof}

\section{Replacement adjacency}\label{app:replacement}
Under replacement adjacency the distinguished element $x$ is replaced by another element $x'$ rather than by $\bot$, and the datasets contain no null elements.
\begin{definition}[Replacement adjacency]\label{def:replacement}
For $n\in\naturals$, an algorithm $\alg$ defined on datasets in $\domain^n$, and $\eps\ge0$, the \emph{replacement privacy profile} is
\[
 \privProfRep{\alg}(\eps)\coloneqq\sup_{\dataset\in\domain^{n-1},\ x,x'\in\domain}\HockeyStickDiv{e^{\eps}}{\alg(\dataset,x)}{\alg(\dataset,x')},
\]
and for an adaptive algorithm $\alg(x,\view)$, the supremum runs over $x,x'\in\domain$ and the view $\view$, comparing $\alg(x,\view)$ with $\alg(x',\view)$. For $\eps\ge0$ and $\delta\in[0,1]$, we say that $\alg$ is \emph{$(\eps,\delta)$-DP under replacement adjacency} if $\privProfRep{\alg}(\eps)\le\delta$. A pair of distributions $(P,Q)$ \emph{dominates $\alg$ under replacement adjacency} if it dominates $(\alg(x,\view),\alg(x',\view))$ for all $x,x'\in\domain$ and every view $\view$.
\end{definition}
Our replacement bound is the smaller of two bounds: a \emph{group bound} that routes $x\to\bot\to x'$ through the remove and add zero-out profiles, and a \emph{neighbor-reference bound} that takes the response of one of the two neighboring elements as the reference distribution. We begin with the group bound.
\begin{lemma}[Group bound]\label{lem:group}
Given $n\in\naturals$ and an algorithm $\alg$, for every $\eps\ge0$,
\begin{equation}\label{eq:triangle}
 \privProfRep{\shuf{n}{\alg}}(\eps)
 \le\min_{u\in[0,\eps]}\Bigl(\privProfRem{\shuf{n}{\alg}}(u)+e^{u}\privProfAdd{\shuf{n}{\alg}}(\eps-u)\Bigr).
\end{equation}
\end{lemma}
\begin{proof}
Let $\dataset\in\domain^{n-1}$, $x,x'\in\domain$ and $u\in[0,\eps]$, and let $A$ be an event. By the variational form $\HockeyStickDiv{\kappa}{P}{Q}=\sup_{A}(P(A)-\kappa Q(A))$, the remove profile at $u$ applied to $(\dataset,x)$ and $(\dataset,\bot)$, and the add profile at $\eps-u$ applied to $(\dataset,x')$ and $(\dataset,\bot)$, give
\begin{align*}
 \shufFunc{n}{\alg}{(\dataset,x)}(A)&\le e^{u}\,\shufFunc{n}{\alg}{(\dataset,\bot)}(A)+\privProfRem{\shuf{n}{\alg}}(u)\\
 &\le e^{\eps}\,\shufFunc{n}{\alg}{(\dataset,x')}(A)+e^{u}\privProfAdd{\shuf{n}{\alg}}(\eps-u)+\privProfRem{\shuf{n}{\alg}}(u).
\end{align*}
Taking the supremum over $A$, $\dataset$, $x$ and $x'$, and the minimum over $u$, gives \cref{eq:triangle}.
\end{proof}
Because the left-hand side of \cref{eq:triangle} does not depend on the response $\alg(\bot,\cdot)$ of the null element, $\alg(\bot,\cdot)$ can be chosen to minimize the bound, with both zero-out profiles on the right-hand side bounded by \cref{thm:reduction}.

The second bound is the following.
\begin{lemma}[Neighbor-reference bound]\label{lem:neighbor-reference}
Given $n\in\naturals$, $\eps_0>0$, $\rep{\delta}_0\in[0,1]$, and an algorithm $\alg$ dominated under replacement adjacency by a pair of distributions $(P,Q)$, if $\privProfRep{\alg}(\eps_0)\le\rep{\delta}_0$, then for every $\eps\ge0$,
\begin{equation}\label{eq:replacement-mixture}
 \privProfRep{\shuf{n}{\alg}}(\eps)
 \le\min\Bigl\{\sum_{k=1}^n w_k\,\privProfRem{\alloc{k}{P,Q}}(\eps),\ \sum_{k=1}^n w_k\,\privProfAdd{\alloc{k}{P,Q}}(\eps)\Bigr\}+(1+e^{\eps})\xi,
\end{equation}
where $p\coloneqq e^{-\eps_0}$, $w_k\coloneqq\prob{}{1+\Bin(n-1,p)=k}$ for $k\in[n]$, and $\xi\coloneqq1-(1-p\rep{\delta}_0)^{n-1}$.
\end{lemma}
\begin{proof}
Fix $x_0\in\domain$ and define $\alg_{x_0}$ by $\alg_{x_0}(x,\view)\coloneqq\alg(x,\view)$ for $x\in\domain$ and $\alg_{x_0}(\bot,\view)\coloneqq\alg(x_0,\view)$ for every view $\view$. Because the output distribution of a shuffled algorithm is determined by the response distributions of its elements, we have $\shufFunc{n}{\alg_{x_0}}{(\dataset,x)}=\shufFunc{n}{\alg}{(\dataset,x)}$ and $\shufFunc{n}{\alg_{x_0}}{(\dataset,\bot)}=\shufFunc{n}{\alg}{(\dataset,x_0)}$ for all $\dataset\in\domain^{n-1}$ and $x\in\domain$. The reference distribution of $\alg_{x_0}$ is $\alg(x_0,\view)$, so by \cref{def:replacement} the pair $(P,Q)$ dominates $\alg_{x_0}$ in the sense of \cref{def:dom} and $\privProfAdd{\alg_{x_0}}(\eps_0)\le\privProfRep{\alg}(\eps_0)\le\rep{\delta}_0$. Applying \cref{thm:reduction} to $\alg_{x_0}$ gives
\begin{align*}
 \HockeyStickDiv{e^{\eps}}{\shufFunc{n}{\alg}{(\dataset,x)}}{\shufFunc{n}{\alg}{(\dataset,x_0)}}&\le\sum_{k=1}^n w_k\,\privProfRem{\alloc{k}{P,Q}}(\eps)+(1+e^{\eps})\xi,\\
 \HockeyStickDiv{e^{\eps}}{\shufFunc{n}{\alg}{(\dataset,x_0)}}{\shufFunc{n}{\alg}{(\dataset,x)}}&\le\sum_{k=1}^n w_k\,\privProfAdd{\alloc{k}{P,Q}}(\eps)+(1+e^{\eps})\xi.
\end{align*}
For $x,x'\in\domain$, the first inequality with $x_0=x'$ and the second with $x$ in place of $x_0$ and $x'$ in place of $x$ bound $\HockeyStickDiv{e^{\eps}}{\shufFunc{n}{\alg}{(\dataset,x)}}{\shufFunc{n}{\alg}{(\dataset,x')}}$ by each of the two sums plus $(1+e^{\eps})\xi$. Taking the supremum over $\dataset$, $x$ and $x'$ gives \cref{eq:replacement-mixture}.
\end{proof}

\para{Comparison.} The bounds are incomparable in general. Taking $u=\eps/2$ in \cref{eq:triangle} shows that the group bound loses about a factor of $2$ in $\eps$ relative to the zero-out bounds, whereas the neighbor-reference bound replaces the zero-out guarantee of $\alg$ by its replacement guarantee in both the dominating pair and the clone probability. For the Gaussian and Laplace mechanisms (as well as binary randomized response), the sensitivity to the null response is half of that between two inputs, so the group bound is lower at every noise level and every $\eps_0\le4$. Conversely, when the null element responds as a fixed input (as for the generic pure DP algorithm of \cref{sec:pure-clones}), the two zero-out bounds from \cref{thm:reduction} coincide with the two sums in \cref{lem:neighbor-reference}. Because these sums are nonincreasing in $\eps$ and $e^{u}\HockeyStickDiv{e^{\eps-u}}{P'}{Q'}\ge\HockeyStickDiv{e^{\eps}}{P'}{Q'}$ for all $P',Q'$ and $u\in[0,\eps]$, the group bound is at least their sum, whereas the neighbor-reference bound equals the smaller of the two.

\begin{figure}[!htbp]
\centering
\includegraphics[width=\linewidth]{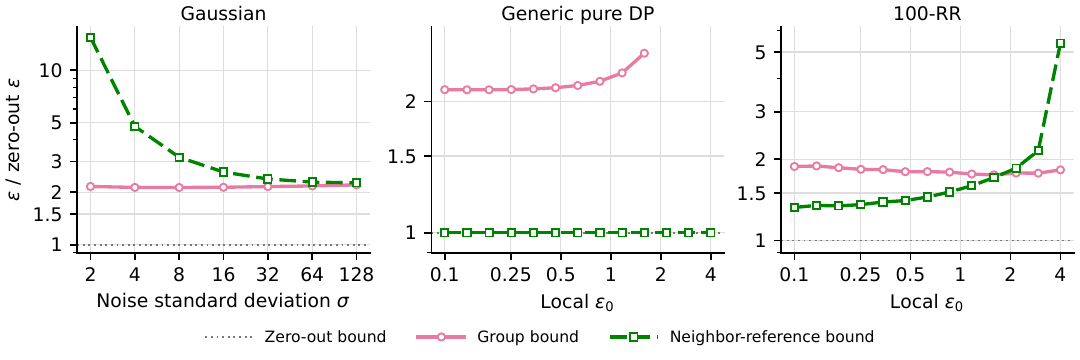}
\caption{The group and neighbor-reference bounds relative to the zero-out bound, for one shuffled pass with $n=1000$ and $\delta=10^{-6}$: the Gaussian mechanism with unit norm bound (left), a generic pure $\eps_0$-DP algorithm, for which the neighbor-reference bound equals the zero-out bound (middle), and $100$-RR with the uniform null response (right). Our replacement bound is the smaller of the two. Points at which a bound reaches the unamplified guarantee are omitted, which removes the group bound of the generic algorithm for $\eps_0\ge\resGenericGroupCapped$. See \cref{app:pure-settings} for the small oscillations in $100$-RR.}
\label{fig:replacement-comparison}
\end{figure}

\Cref{fig:replacement-comparison} illustrates both extremes together with an intermediate case: $k$-ary randomized response with $k=100$, which outputs its input with probability $e^{\eps_0}/(e^{\eps_0}+k-1)$ and every other category with probability $1/(e^{\eps_0}+k-1)$, and whose null response is uniform. For large $k$, the zero-out $\eps$ of a category against the uniform response, $\ln\bigl(ke^{\eps_0}/(e^{\eps_0}+k-1)\bigr)$, is close to $\eps_0$, so at small $\eps_0$ the zero-out pair does not compensate for the factor of $2$ in the group bound. Thus for $100$-RR the neighbor-reference bound is lower for $\eps_0$ below about $\resHundredRRCross$, whereas the group bound is lower above it, where the zero-out clone probability $k/(e^{\eps_0}+k-1)$ is much larger than $e^{-\eps_0}$.

\section{Proof for \texorpdfstring{\cref{sec:gauss-shuffle}}{Section 4}}\label{app:gauss-shuffle-proofs}
In this appendix we give the full proof of \cref{thm:gauss-shuffle}, writing $\Phibar\coloneqq1-\Phi$ for the standard normal tail.

\begin{proofof}{\cref{thm:gauss-shuffle}}
Let $(P,Q)\coloneqq(\Nrm(1,\sigma^2),\Nrm(0,\sigma^2))$ be the Gaussian pair. By \cref{clm:gauss-dom}, \cref{eq:local-simulation}, post-processing (\cref{lem:hs-facts}), and the symmetric Gaussian hockey-stick formula~\cref{eq:gauss-profile}, for every $x$, $\view$, and $\eps_0>0$,
\[
 \HockeyStickDiv{e^{\eps_0}}{Q_{\view}}{\GaussMech{\sigma}(x,\view)}\le\HockeyStickDiv{e^{\eps_0}}{Q}{P}
 =\Phi\left(\frac1{2\sigma}-\eps_0\sigma\right)-e^{\eps_0}\Phi\left(-\frac1{2\sigma}-\eps_0\sigma\right)
 \le\Phibar\left(\eps_0\sigma-\frac1{2\sigma}\right).
\]
Set $\eps_0\coloneqq1/(2\sigma^2)+\sqrt{2\ln(16n/\delta)}/\sigma$ and $p\coloneqq e^{-\eps_0}$, so that the Gaussian tail bound $\Phibar(z)\le e^{-z^2/2}$ gives $\privProfAdd{\GaussMech{\sigma}}(\eps_0)\le\delta_0\coloneqq\delta/(16n)$. In both claims $\sigma\ge3\sqrt{\ln(n/\delta)}$: in the first we take $C_1\ge3$, and in the second it holds for all sufficiently large $n$, since there $\ln(n/\delta)=\ln(n)+\ln(1/\delta)=(1+o(1))\ln(1/\delta)$ and $\sigma=\omega\bigl(\sqrt{\ln(1/\delta)}\bigr)$. As $\ln(n/\delta)\ge\ln(4)$, we have that $\ln(16n/\delta)\le3\ln(n/\delta)$, and therefore $\eps_0\le1/(18\ln(4))+\sqrt6/3<1$ and $p>1/e$. Let $m\coloneqq\max\{1,\lfloor pn-\sqrt{2pn\ln(4/\delta)}\rfloor\}$. For $\eps\le1$ the last term of the single-count bound~\cref{eq:cutoff} is at most $(1+e)\delta/16\le\delta/4$, and therefore \cref{thm:reduction} with $\beta=\delta/4$ gives, for every $\eps\in[0,1]$,
\[
 \privProfRem{\shuf{n}{\GaussMech{\sigma}}}(\eps)\le\privProfRem{\alloc{m}{P,Q}}(\eps)+\frac\delta2
 \qquad\text{and}\qquad
 \privProfAdd{\shuf{n}{\GaussMech{\sigma}}}(\eps)\le\privProfAdd{\alloc{m}{P,Q}}(\eps)+\frac\delta2 .
\]
The pairs $(\Pbar{m},Q^m)$ and $(Q^m,\Pbar{m})$ are the pairs of output distributions of $1$-out-of-$m$ random allocation of the one-dimensional Gaussian mechanism with $f\equiv1$. As the privacy profiles are nonincreasing in $\eps$, it therefore suffices to show that this algorithm is $(\eps,\delta/2)$-DP for the claimed $\eps$, and that $\eps\le1$.

For the first claim, let $C_0\ge1$ be the constant in \citep[Corollary~5.4]{FS26clt}, and set $C_1=C\coloneqq2\sqrt3\,C_0>3$ and $C_2\coloneqq96C_0^3$. Since $\ln(n/\delta)\ge\ln(4)$ and $\ln(4/\delta)\le3\ln(1/\delta)$, the condition $n\ge C_2\ln^2(n/\delta)$ implies $n\ge96\ln^2(4)>60$ and $pn\ge n/e\ge8\ln(4/\delta)$, so $\sqrt{2pn\ln(4/\delta)}\le pn/2$ and $m\ge n/(2e)-1\ge n/6\ge10$. We apply \citep[Corollary~5.4]{FS26clt} with $t=m$ and $\delta/2$ in place of $\delta$ (since $\sigma\ge3\sqrt{\ln(4)}>1$, $m\ge10$ and $\delta/2\le1/4$). Its hypothesis requires $\sigma\ln(1+a/\sigma)\ge C_0\sqrt{\ln(2m/\delta)}$, where $a\coloneqq\sqrt{m/(C_0\ln(2/\delta))}$; we split into two cases according to whether $a/\sigma\ge1$. If $a/\sigma\ge1$, then $\ln(1+a/\sigma)\ge\ln(2)>1/2$, so $\sigma\ln(1+a/\sigma)\ge\frac12 C_1\sqrt{\ln(n/\delta)}\ge C_0\sqrt{\ln(2m/\delta)}$ since $\ln(2m/\delta)\le2\ln(n/\delta)$. If $a/\sigma<1$, then $\ln(1+a/\sigma)\ge a/(2\sigma)$, and using $m\ge n/6$, $\ln(2/\delta)\le2\ln(1/\delta)$ and $n\ge96C_0^3\ln(1/\delta)\ln(n/\delta)$ gives $\sigma\ln(1+a/\sigma)\ge a/2\ge C_0\sqrt{\ln(2m/\delta)}$. Thus the condition of \citep[Corollary~5.4]{FS26clt} holds in both cases, showing that $\alloc{m}{P,Q}$ is $(\eps_1,\delta/2)$-DP in both directions for $\eps_1=C_0\sqrt{\ln(2/\delta)}/(\sigma\sqrt m)\le C\sqrt{\ln(1/\delta)}/(\sigma\sqrt n)=\eps$. Finally, $\sigma\ge C_1\sqrt{\ln(n/\delta)}\ge C\sqrt{\ln(1/\delta)}$ gives $\eps\le1/\sqrt n<1$.

For the second claim, let $L\coloneqq\ln(1/\delta)$. Since $L=\omega(\ln(n))$, each of $\ln(2/\delta)$, $\ln(4/\delta)$ and $\ln(16n/\delta)$ is $L+O(\ln(n))=(1+o(1))L$. Hence $\sigma=\omega(\sqrt L)$ gives $\sigma\to\infty$ and $\ln(16n/\delta)/\sigma^2\to0$, so that $\eps_0\to0$ and $p\to1$, and $L=o(\sqrt n)$ gives $\ln(4/\delta)=o(n)$, so that also $m=(1-o(1))n$. We apply \citep[Corollary~E.2]{FS26clt} with $t=m$ and $\delta/2$ in place of $\delta$, under the second of its two sufficient conditions: $\sigma$ is bounded away from zero, $\ln(2/\delta)=\omega(\ln(m))$, $\ln(2/\delta)=o(\sqrt m)$, and $\sigma=\Omega\bigl(\sqrt{\ln(2/\delta)}\bigr)$. The first holds since $\sigma\to\infty$, the second since $\ln(2/\delta)\ge L=\omega(\ln(n))$ and $m\le n$, the third since $\ln(2/\delta)=(1+o(1))L=o(\sqrt n)$ and $m=(1-o(1))n$, and the fourth since $\ln(2/\delta)=(1+o(1))L$ and $\sigma=\omega(\sqrt L)$. That corollary allows $\sigma$ and $\delta$ to depend on $t$, and its proof uses only these limits, so it applies along $t=m$ although $\sigma$ and $\delta$ are functions of $n$. It gives that $\alloc{m}{P,Q}$ is $(\eps_m,\delta/2)$-DP in both directions for
\[
 \eps_m=(1+o(1))\sqrt{\frac{2\left(e^{1/\sigma^2}-1\right)\ln(2/\delta)}{m}}=(1+o(1))\frac{\sqrt{2\ln(1/\delta)}}{\sigma\sqrt n},
\]
since $m=(1-o(1))n$, $\ln(2/\delta)=(1+o(1))\ln(1/\delta)$ and $e^{1/\sigma^2}-1=(1+o(1))/\sigma^2$. In particular, $\eps_m\le1$ for all sufficiently large $n$.
\end{proofof}

\section{\texorpdfstring{Proofs and details for \cref{sec:lower-bound}}{Proofs and details for Section 5}}\label{app:lower-bound}
In this appendix we prove \cref{lem:identical-ratio,lem:pool} and describe the nonnegative score-sum tests and multi-pass PLD lower-bound computation used in \cref{sec:lower-bound,sec:dpsgd}.

\begin{proofof}{\cref{lem:identical-ratio} (Single-pass likelihood-ratio test)}
For every $z\in\reals^n$, expanding the densities $U_x(z)=\frac1n\sum_{j=1}^nP(z_j)\prod_{l\ne j}G(z_l)$ and $U_\bot(z)=\frac1n\sum_{j=1}^nQ(z_j)\prod_{l\ne j}G(z_l)$ gives
\[
 U_x(z)-\kappa U_\bot(z)=\frac1n\prod_{l=1}^nG(z_l)\cdot\sum_{j=1}^n\Delta_\kappa(z_j).
\]
Thus the integrand $[U_x(z)-\kappa U_\bot(z)]_+$ in the definition of $\HockeyStickDiv{\kappa}{U_x}{U_\bot}$ equals $U_x(z)-\kappa U_\bot(z)$ when $\sum_j\Delta_\kappa(z_j)>0$ and $0$ otherwise, and integrating it gives \cref{eq:identical-hs}. The sum $\sum_j\Delta_\kappa(z_j)$ does not depend on the order of the responses, so under $U_x$ it has the same distribution as when the distinguished response is in the last position, that is, when $y_1,\ldots,y_{n-1}\sim G$ and $y_n\sim P$ are independent. The same holds under $U_\bot$ with $Q$ in place of $P$.
\end{proofof}

\begin{proofof}{\cref{lem:pool} (Dominated pair from test bounds)}
Let $(0,0)=(\hat q_0,\hat p_0),\ldots,(\hat q_s,\allowbreak\hat p_s)=(1,1)$ be the vertices of the upper concave hull of $\mathcal T\coloneqq\{(0,0),(1,1)\}\cup\{(q_i,p_i):i\in[r]\}$, ordered by strictly increasing first coordinate, and define $\widetilde P(j)\coloneqq\hat p_j-\hat p_{j-1}$ and $\widetilde Q(j)\coloneqq\hat q_j-\hat q_{j-1}$ for $j\in[s]$. Since $p_i=0$ whenever $q_i=0$ and $p_i\le1$ for all $i\in[r]$, the hull starts at $(0,0)$ and ends at $(1,1)$, and every vertex belongs to $\mathcal T$, so $s\le r+1$. From $0=\hat q_0<\dots<\hat q_s=1$ we obtain $\widetilde Q(j)>0$ for all $j\in[s]$ and $\sum_{j=1}^s\widetilde Q(j)=1$. Because $\hat p_{s-1}\le1$, the last hull slope $(1-\hat p_{s-1})/(1-\hat q_{s-1})$ is nonnegative, and by concavity the slopes $\widetilde P(j)/\widetilde Q(j)$ are nonincreasing in $j$. Thus $\widetilde P(j)\ge0$ for all $j\in[s]$ and $\sum_{j=1}^s\widetilde P(j)=1$. Finally, for every $\kappa\ge0$, the indices $j\in[s]$ with $\widetilde P(j)>\kappa\widetilde Q(j)$ form a prefix, giving
\[
 \HockeyStickDiv{\kappa}{\widetilde P}{\widetilde Q}
 =\max_{0\le j\le s}(\hat p_j-\kappa\hat q_j)
 =\max_{(q,p)\in\mathcal T}(p-\kappa q)
 \le\HockeyStickDiv{\kappa}{R}{S}.
\]
Here the second equality holds because every vertex lies in $\mathcal T$ and every point in $\mathcal T$ lies on or below the concave hull (so $p-\kappa q$ is bounded by a convex combination of values at consecutive vertices), and the inequality follows from $p_i-\kappa q_i\le R(A_i)-\kappa S(A_i)\le\HockeyStickDiv{\kappa}{R}{S}$ for every $i\in[r]$, together with the empty and full events at $(0,0)$ and $(1,1)$.
\end{proofof}

\para{Nonnegative score-sum tests.}
Focusing on the remove direction $(U_x,U_\bot)$ (which suffices for a valid DP lower bound), computing the test probabilities in \cref{lem:identical-ratio} via an FFT on a uniform grid works for a single threshold, since each term can be truncated above the value at which it makes the sum positive on its own (\cref{app:implementation}). The multi-pass bound below, however, needs the tail probabilities at all thresholds, where no such truncation is possible. At low noise the untruncated terms $\Delta_\kappa(y)=e^{2y/\sigma^2}-\kappa e^{(y+1/2)/\sigma^2}$ span too many orders of magnitude for a uniform grid. Multiscale FFT (MS-FFT)~\citep{FS26msfft} computes convolutions on dyadic grids across arbitrary dynamic ranges, but only for nonnegative terms, whereas $\Delta_\kappa$ takes negative values whenever $\kappa>0$. A shift by a common constant makes all terms nonnegative, and this step relies on the Gaussian densities: $P/G(y)=e^{2y/\sigma^2}$ is a constant multiple of the square of $Q/G(y)=e^{(y+1/2)/\sigma^2}$, so $\Delta_\kappa$ is a quadratic in $e^{y/\sigma^2}$, and completing the square with $\beta_\kappa\coloneqq\frac{\kappa}{2}e^{1/(2\sigma^2)}$ gives
\[
 X_\kappa(y)\coloneqq\Delta_\kappa(y)+\beta_\kappa^2=\bigl(e^{y/\sigma^2}-\beta_\kappa\bigr)^2\ge0
 \qquad\text{and}\qquad
 S_\kappa(z)\coloneqq\sum_{j=1}^n X_\kappa(z_j).
\]
We use the tests $\{S_\kappa\ge g\}$ for $\kappa,g\ge0$. For $g=n\beta_\kappa^2$ this is the likelihood-ratio test of \cref{lem:identical-ratio} (up to a null set), and for $\kappa=0$ it thresholds the numerator of the likelihood ratio in \cref{eq:identical-ratio}.

By \cref{lem:identical-ratio}, under $U_x$ the score $S_\kappa(z)$ is the sum of $n-1$ independent copies of $X_\kappa(y)$ with $y\sim G$ and one with $y\sim P$, and under $U_\bot$ the last one has $y\sim Q$. We discretize each term onto a dyadic grid (with two unbounded end cells) and convolve with MS-FFT, rounding the terms and the partial sums down in the computation for $U_x$ and up in the computation for $U_\bot$. For each $\kappa$, the two runs give at every grid threshold $g$ a lower bound $p_{\kappa,g}\le U_x(S_\kappa\ge g)$ and an upper bound $q_{\kappa,g}\ge U_\bot(S_\kappa\ge g)$.

\para{Multi-pass PLD lower bound.}
Applying the construction of \cref{sec:lower-bound} (\cref{lem:pool}) with the events $\{S_\kappa\ge g\}$ and bounds $(p_{\kappa,g},q_{\kappa,g})$ over any finite collection of pairs $(\kappa,g)$ (with $q_{\kappa,g}>0$) yields the dominated pair $(\widetilde P,\widetilde Q)$ whose $E$-fold PLD convolution bounds $(U_x^E,U_\bot^E)$. We compute this convolution $\mu_E$ by repeated squaring on a grid $h\mathbb Z$ with step $h>0$, rounding the losses $\ln(\widetilde P(j)/\widetilde Q(j))$ down to $h\mathbb Z$ and discarding small tails without renormalizing. Both steps can only decrease the expectation of the nonnegative nondecreasing function $\loss\mapsto[1-e^{\eps-\loss}]_+$, and therefore, for every $\eps\ge0$,
\begin{equation}\label{eq:multipass-lb}
 \underline H(\eps)\coloneqq\sum_{\loss>\eps}\mu_E(\loss)\bigl(1-e^{\eps-\loss}\bigr)
 \le\HockeyStickDiv{e^\eps}{\widetilde P^E}{\widetilde Q^E}
 \le\HockeyStickDiv{e^\eps}{U_x^E}{U_\bot^E}.
\end{equation}
Any $\underline\eps>0$ with $\underline H(\underline\eps)>\delta$ thus shows that $E$ shuffled passes of $\GaussMech{\sigma}$ with independent permutations fail to be $(\eps,\delta)$-DP under zero-out adjacency for every $\eps<\underline\eps$. The additional thresholds $g\ne n\beta_\kappa^2$ returned by the same MS-FFT run can only tighten \cref{eq:multipass-lb}. We describe the implementation details in \cref{app:implementation}.

\section{Details of the experiments}\label{app:experiments}
In this appendix we describe the computations behind the curves of \cref{sec:evaluation} that are specific to this work. We cap our upper bounds at the unamplified guarantee of the algorithm, which also holds for the shuffled algorithm (\cref{sec:statement}).

\subsection{Implementation}\label{app:implementation}
We compute the allocation PLDs with the \texttt{PLD\_accounting} library \citep{PLD_acc}, and the PLD of $t$-step Poisson subsampling of the Gaussian pair $(P,Q)$ with sampling probability $1/t$ (the $t$-fold composition of the PLD of $((1-1/t)Q+(1/t)P,Q)$) together with its optimistic numerical bracket using Google's accounting library \citep{Google20}. Except for the multi-pass lower bound below, all computations use floating-point arithmetic and are therefore not interval-arithmetic certificates, although every discretization errs toward a larger upper bound and a smaller lower bound.

For the clones baseline in~\citep[Theorem~3.8]{FMT21} in \cref{fig:gauss-replacement}, we split $\delta$ between their bound for pure $\eps_0$-DP algorithms and the additional term $(e^{\eps}+1)(1+e^{-\eps_0}/2)n\delta_0$, take $\eps_0$ from the pair $(\Nrm(2,\sigma^2),\Nrm(0,\sigma^2))$ at the resulting $\delta_0$, and report the smallest bound over five splits, capped at the unamplified guarantee (which it reaches for $\sigma\le\resClonesCapSigma$).

\para{Grouping the counts.}
As shown in the proof of \cref{thm:reduction}, the allocation profiles are nonincreasing in the count. Instead of computing them at every count, we therefore partition the counts into intervals at binomial quantiles of $K=1+\Bin(n-1,p)$ and bound the profile on each interval by its value at the left endpoint, weighted by the binomial probability of the interval. The profile at the right endpoint of each interval gives a lower estimate of the count mixture, so that refining the partition brackets its exact value. The only failure probability spent on the binomial tail is the negligible probability of the counts below the first interval, which we add to $\delta$. For the single-count bound \cref{eq:cutoff}, we retain only the allocation PLD at count $m$, with weight $\prob{}{K\ge m}$, and assign the remaining mass $\prob{}{K<m}$ to a disjoint pair before applying the same coupling correction and envelope construction. For each clone probability, we take the largest $m$ whose lower count-tail probability is at most $0.05\delta$, $0.2\delta$, $0.5\delta$, or $0.9\delta$, and combine the resulting four pairs. The shaded brackets in \cref{fig:comparison} use the lower estimates of the random-allocation profile from \citep{FS26msfft} and the optimistic Poisson accountant.

\para{Combining clone probabilities.}
When the discrete allocation PLDs share a finite loss grid $\{\loss_i\}$, we can construct the envelope pair $(P_*,Q_*)$ of \cref{sec:pld-accounting} for all clone probabilities $p\in\mathcal P$ at once. Specifically, in the remove direction we cache the mass arrays for the selected count-interval endpoints (padding with zeros to align grid indices and placing any omitted lower count-tail probability at $+\infty$) and form each mixture PLD $\mu_p$ by multiplying its vector of grouped binomial weights with the cached allocation mass arrays. The linear-time tail transformation of \cref{lem:pld-tv} removes an upper tail of denominator mass $\xi_p$ and a lower tail of numerator mass $\xi_p$, moving the removed upper-tail mass plus $\xi_p$ to $+\infty$ (\cref{app:reduction-proofs}) and preserving the finite loss grid $\{\loss_i\}$. At each grid loss $\loss_i$, the hockey-stick divergence of the corrected PLD $\widehat\mu_p$ is
\[
 d_{p,i}=\widehat\mu_p(+\infty)+\sum_{j>i}\widehat\mu_p(\loss_j)\bigl(1-e^{\loss_i-\loss_j}\bigr),
\]
which two suffix sums evaluate at all knots in linear time in the grid size. Taking $d_i\coloneqq\min_{p\in\mathcal P}d_{p,i}$ at each knot $\kappa_i\coloneqq e^{\loss_i}$ yields the tightest remove-direction cloning bound at every grid threshold, so no separate threshold grid is needed for the cloning terms. We perform the analogous calculation with the add-direction allocation PLDs and convert each add-direction knot bound $d^{\mathrm{add}}$ at $1/\kappa$ into a remove-direction bound $1-\kappa+\kappa d^{\mathrm{add}}$ at $\kappa$ (via the complement identity $\HockeyStickDiv{\kappa}{P'}{Q'}=1-\kappa+\kappa\HockeyStickDiv{1/\kappa}{Q'}{P'}$ and \cref{lem:kov}). In exact arithmetic these bounds coincide with the remove-direction ones, but since the add-direction PLDs are discretized separately, they tighten the hull for $\kappa<1$. We combine all resulting remove-direction points with $(0,1)$ and sampled hockey-stick bounds of the unamplified pair $(P,Q)$ (at both $\kappa$ and $1/\kappa$, ensuring each bound is at least $[1-\kappa]_+$), and compute their nonincreasing lower convex hull $H_*$ in exact rational arithmetic. In the resulting piecewise-linear curve $H_*$, each positive vertex $\kappa_j$ where the slope jumps from $-s_{j-1}$ to $-s_j$ (with $\Delta s_j\coloneqq s_{j-1}-s_j>0$) contributes masses $(\kappa_j\Delta s_j,\Delta s_j)$ to $(P_*,Q_*)$. We add an atom of mass $1-s_0$ at ratio $0$ under $Q_*$ only and an atom of mass $\lim_{\kappa\to\infty}H_*(\kappa)$ at $+\infty$ under $P_*$ only, then query the resulting finite PLD directly for a single pass. For multiple passes we apply upward loss rounding and compose its remove and add PLDs in Google's accounting library \citep{Google20}. See \cref{app:dpsgd} for the grid parameters and multi-clone ablation used in \cref{sec:dpsgd}.

For the single-pass Gaussian curves, we use $1001$ logarithmically spaced clone probabilities from $\max\{10^{-9},10^{-5}p_{\max}\}$ to $p_{\max}$, where $2\xi_{p_{\max}}=\HockeyStickDiv{1}{P}{Q}$, and discard probabilities whose $2\xi_p$ is already at least the best total-variation bound. We use a loss step of $0.005$ for the sampled unamplified Gaussian bounds and query the envelope without further loss rounding, also capping the result by the exact unamplified guarantee. The allocation PLDs use $\min\{2^{16},\max\{2^{12},2^{\lceil\log_2(2048\sigma)\rceil}\}\}$ cells per octave, with $\sigma/2$ in place of $\sigma$ for the neighbor-reference bound. Count intervals have relative width at most $0.05$; the allocation tail tolerance is $10^{-14}$ and the hockey-stick bounds receive an additive padding of $2\times10^{-14}$. The group bound uses the two profiles of the zero-out envelope, and the neighbor-reference bound uses the smaller directional bound at noise scale $\sigma/2$. For the zero-out and single-count curves at $n=1000$ and $\sigma\in\{1,8,128\}$, and the zero-out curve at $n=10000$ and $\sigma=1$, doubling the clone-probability grid to $2001$ points changes $\eps$ by less than $\resCloneGridChange\%$. Doubling the allocation resolution at $n=1000$ and $\sigma\in\{1,8\}$ changes the count-mixture bounds by less than $\resCellsChange\%$.

\para{Lower-bound accounting.}
For the single-pass lower bound in \cref{fig:comparison} we bound $U_x(A)$ from below and $U_\bot(A)$ from above for a rounded version $A$ of the event $\{\sum_j\Delta_\kappa(z_j)>0\}$ (\cref{lem:identical-ratio}). In this event, $\Delta_\kappa$ is replaced on each of $400{,}000$ cells by its minimum over the cell, rounded down to a lattice and truncated far above the value at which a single term can make the sum positive. Since every event gives a valid lower bound, the rounding affects only how far the test is from the optimal one. We compute the masses of the rounded terms from Gaussian cell probabilities, rounded down for $U_x(A)$ and up for $U_\bot(A)$, and the distribution of the sum of the $n-1$ common terms by FFT convolutions. The mass that leaves the window of each partial sum is dropped or moved so that it can only decrease the lower bound and increase the upper bound, and we subtract an explicit allowance for the floating-point error of the FFT convolutions.

For the multi-pass lower bound of \cref{app:lower-bound} used in \cref{sec:dpsgd}, we use the events $\{S_\kappa\ge g\}$ at all grid thresholds $g$ for $\kappa\in\{0,1\}$ when $\sigma<1$ and $\kappa\in\{1,1.25,1.5\}$ when $\sigma\ge1$. Using more values of $\kappa$ did not lead to appreciable gains. We discretize each $X_\kappa$ with $2^{16}$ cells per octave in \texttt{PLD\_accounting} \citep{PLD_acc}, covering $\pm12\sigma$ around the means of $G$, $P$ and $Q$, and we round the losses of $(\widetilde P,\widetilde Q)$ down onto $h\mathbb Z$ with step $h=10^{-4}$ ($h=2\times10^{-4}$ for $\sigma<0.45$), cropping tails of mass at most $10^{-22}$ without renormalizing during the $E$-fold convolution. Every numerical step of this multi-pass lower bound is enclosed outward. We bound Gaussian cell probabilities, log-ratios, and the final check $\underline H(\underline\eps)>\delta$ in $128$-bit ball arithmetic using Arb \citep{Johansson17Arb}, and compute polynomial products inside MS-FFT (after outward quantization to $2^{-160}\mathbb Z$) and in $\mu_E$ exactly over the integers. We bound nonnegative sums with standard floating-point summation error bounds, and compute the upper concave hull in exact rational arithmetic.

\subsection{Pure DP algorithms}\label{app:pure-settings}
For binary randomized response with uniform null response $U=\frac12(P_0+P_1)$, replacement privacy implies zero-out privacy by joint convexity, which follows from post-processing and revealed mixtures (\cref{lem:hs-facts}). For the Laplace mechanism in \cref{sec:pure-clones}, the stated dominating pairs hold since for $\eps\ge0$ the hockey-stick divergence $\HockeyStickDiv{e^{\eps}}{\cdot}{\cdot}$ between the responses to two inputs at distance $d$ is $\bigl(1-\exp((\eps-\eps_0d)/2)\bigr)_+$ in both directions, which increases with $d$, and since $\HockeyStickDiv{\kappa}{P'}{Q'}=1-\kappa+\kappa\HockeyStickDiv{1/\kappa}{Q'}{P'}$ for $\kappa<1$ (by taking complements in the variational form $\sup_A(P'(A)-\kappa Q'(A))$).

For the plotted randomized-response bounds, the \texttt{PLD\_accounting} library computes the allocation PLDs by MS-FFT using an upper discretization of each pair, with clone counts grouped as in \cref{app:implementation}. In \cref{fig:replacement-comparison}, the ratios for $100$-RR oscillate by about $1\%$ because its privacy loss takes few distinct values in the tail. By \citep[Lemma~5.1]{FS26clt}, the likelihood ratio of $(\Pbar{m},Q^m)$ under $Q^m$ is the average of $m$ independent copies of $P/Q$ under $Q$, which for randomized response depends only on the binomial count (a trinomial count for $k$-RR under replacement adjacency) of outputs in the distinguished category. An exact evaluation via this count distribution oscillates in the same way. We compute the baselines in~\citep{FMT21} with their numerical accountant, and those in~\citep{FMT23} by evaluating the four-label experiment in~\citep[Lemma~7.3]{FMT23} exactly. For randomized response we use the decomposition in~\citep[Theorem~7.4]{FMT23}, which gives the lowest bound over all decompositions. For the Laplace mechanism we decompose the two neighboring densities $A$ and $B$ into components proportional to $(A-B)_+$, $(B-A)_+$ and $e^{\eps_0}\min(A,B)-\max(A,B)$, and we take the worst case over all pairs of neighboring inputs in $[0,1]$. A lower bound over all decompositions shows that the resulting $\eps$ is within $0.2\%$ of the lowest one.

\section{Published DP-SGD settings}\label{app:dpsgd}
\para{Settings.}
Each setting in \cref{tab:dpsgd-settings} is one training run from one of the seven papers of \cref{sec:dpsgd}. We use only settings with at least $20$ full batches per epoch. We take the parameters of each setting from the paper and, where the paper does not give them, from its released code.

\begin{sidewaystable}
\centering\footnotesize\setlength{\tabcolsep}{3.2pt}
\begin{tabular}{@{}rl l rrrrrrr c l r r r r r@{}}
\toprule
 & & & & & & & \multicolumn{2}{c}{Epochs} & & & & \multicolumn{3}{c}{Reported} & \multicolumn{2}{c}{Poisson PLD}\\
\cmidrule(lr){8-9}\cmidrule(lr){13-15}\cmidrule(l){16-17}
ID & Ref. & Dataset & $N$ & $B$ & $q$ & $T$ & stated & $Tq$ & $C$ & Sampler & Accountant & $\delta$ & $\eps$ & $\sigma$ & $\sigma$ & change\\
\midrule
1 & \cite{TB21} & Fashion-MNIST & 60,000 & 2,048 & 0.0341 & 1,160 & 40 & 39.6 & 0.1 & S & RDP & $10^{-5}$ & 3 & 2.15 & 1.81 & ${-}16.2\%$ \\
2 & \cite{TB21} & CIFAR-10 & 50,000 & 2,048 & 0.0410 & 960 & 40 & 39.3 & 0.1 & S & RDP & $10^{-5}$ & 2 & 3.32 & 2.68 & ${-}19.4\%$ \\
3 & \cite{TB21} & CIFAR-10 & 50,000 & 1,024 & 0.0205 & 1,440 & 30 & 29.5 & 0.1 & S & RDP & $10^{-5}$ & 3 & 1.54 & 1.32 & ${-}14.7\%$ \\
\addlinespace[3pt]
4 & \cite{BPS19} & DiF + UTKFace & 30,000 & 256 & 0.00853 & 7,032 & 60 & 60.0 & 1 & S & RDP & $10^{-6}$ & 9.16 & 0.8 & 0.749 & ${-}6.4\%$ \\
5 & \cite{BPS19} & iNaturalist-60k & 60,000 & 32 & 0.000533 & 56,250 & 30 & 30.0 & 1 & S & RDP & $10^{-6}$ & 4.67 & 0.6 & 0.546 & ${-}9.1\%$ \\
\addlinespace[3pt]
6 & \cite{BDSHSSKSB23} & CheXpert & 223,414 & 4,096 & 0.0183 & 3,000 & -- & 55.0 & 0.001 & U & PLD & $4.48\cdot10^{-6}$ & 8 & 0.91 & 0.912 & $+0.2\%$ \\
7 & \cite{BDSHSSKSB23} & MIMIC-CXR & 259,000$^*$ & 4,096 & 0.0158 & 3,000 & -- & 47.4 & 0.001 & U & PLD & $3.86\cdot10^{-6}$ & 8 & 0.85 & 0.848 & ${-}0.3\%$ \\
\addlinespace[3pt]
8 & \cite{YNBGIKKLMWYZ22} & SST-2 & 67,349 & 2,000 & 0.0297 & 660 & 20 & 19.6 & 10 & S & PRV & $10^{-5}$ & 6.7 & 0.92 & 0.862 & ${-}6.4\%$ \\
9 & \cite{YNBGIKKLMWYZ22} & QNLI & 104,743 & 2,000 & 0.0191 & 1,040 & 20 & 19.9 & 10 & S & PRV & $10^{-5}$ & 6.7 & 0.83 & 0.778 & ${-}6.4\%$ \\
10 & \cite{YNBGIKKLMWYZ22} & MNLI & 392,702 & 2,000 & 0.00509 & 3,920 & 20 & 20.0 & 10 & S & PRV & $2.55\cdot10^{-9}$ & 1.61 & 1.88 & 1.32 & ${-}29.9\%$ \\
11 & \cite{YNBGIKKLMWYZ22} & E2E NLG & 42,061 & 128 & 0.00304 & 6,560 & 20 & 20.0 & 1 & S & PRV & $10^{-5}$ & 6 & 0.6 & 0.584 & ${-}2.8\%$ \\
12 & \cite{YNBGIKKLMWYZ22} & DART v1.1.1 & 62,659 & 256 & 0.00409 & 4,880 & 20 & 19.9 & 1 & S & PRV & $10^{-5}$ & 6.8 & 0.6 & 0.588 & ${-}2.2\%$ \\
\addlinespace[3pt]
13 & \cite{LZLRMS23} & IMDb & 25,000 & 64 & 0.00256 & 39,000 & 100 & 99.8 & 0.5 & S & RDP & $10^{-5}$ & 3.04 & 1 & 0.955 & ${-}4.5\%$ \\
14 & \cite{LZLRMS23} & StackOverflow & 246,092 & 64 & 0.000260 & 192,250 & 50 & 50.0 & 0.25 & S & RDP & $10^{-6}$ & 0.87 & 1 & 0.834 & ${-}16.6\%$ \\
\addlinespace[3pt]
15 & \cite{BDLS20} & MNIST & 60,000 & 256 & 0.00427 & 3,516 & 15 & 15.0 & 1.5 & U & GDP/CLT & $10^{-5}$ & 0.83 & 1.3 & 1.34 & $+2.7\%$ \\
16 & \cite{BDLS20} & Adult income & 29,305 & 256 & 0.00874 & 2,052 & 18 & 17.9 & 1 & P & GDP/CLT & $10^{-5}$ & 10.2 & 0.55 & 0.575 & $+4.5\%$ \\
17 & \cite{BDLS20} & IMDb & 25,000 & 512 & 0.0205 & 432 & 9 & 8.85 & 1 & P & GDP/CLT & $10^{-5}$ & 10.43 & 0.56 & 0.589 & $+5.1\%$ \\
18 & \cite{BDLS20} & MovieLens 1M & -- & -- & 0.0125 & 1,600 & 20 & 20.0 & 5 & U & GDP/CLT & $10^{-6}$ & 10.61 & 0.6 & 0.638 & $+6.3\%$ \\
\addlinespace[3pt]
19 & \cite{KSCGTT22} & ImageNet-1k & 1,281,167 & 1,024 & 0.000799 & 12,511 & 10 & 10.0 & 1 & S & RDP & $10^{-6}$ & 4.6 & 0.56 & 0.533 & ${-}5.0\%$ \\
20 & \cite{KSCGTT22} & ImageNet-1k & 1,281,167 & 1,024 & 0.000799 & 12,511 & 10 & 10.0 & 1 & S & RDP & $10^{-6}$ & 71 & 0.28 & 0.267 & ${-}5.0\%$ \\
\bottomrule
\end{tabular}

\caption{The DP-SGD settings of \cref{sec:dpsgd}: number $N$ of training examples, batch size $B$, sampling probability $q$, number of steps $T$, stated and expected ($Tq$) number of epochs, clipping norm $C$, training sampler of the released code (S: shuffled or fixed batches, P: Poisson sampling, U: not verified), the accountant of the reported guarantee (RDP: R\'enyi DP; PRV: the privacy random variable accountant \citep{GLW21}; GDP/CLT: the central limit approximation of Gaussian DP \citep{DRS22}), and the reported $\delta$, $\eps$ and $\sigma$. The last two columns give the $\sigma$ for which the PLD accountant for Poisson subsampling achieves the reported $(\eps,\delta)$ with the given $q$ and $T$, and its percentage change relative to the reported $\sigma$. We round the PLD noise multiplier up and compute the change before rounding. $^*$Approximate $N$.}
\label{tab:dpsgd-settings}
\end{sidewaystable}

\para{Shuffled runs and matching Poisson subsampling.}
In the shuffled model of \cref{sec:dpsgd}, each of the $E$ epochs splits a fresh uniformly random permutation of $nB$ examples into $n$ batches of size $B$ ($n=80$ for MovieLens, where the paper reports $q=1/80$ rather than $B$), with $E=\lceil T/n\rceil$ so that the last epoch is complete. Running Poisson subsampling on these $nB$ examples with sampling probability $1/n$ over $En$ steps matches both the expected batch size and the number of steps of the shuffled run; its noise multiplier agrees with the recomputed Poisson $\sigma$ for the original $q$ and $T$ to within $\resMatchChange\%$ (\cref{tab:dpsgd-noise}). For Poisson subsampling, zero-out adjacency gives the same pairs of output distributions as the add and remove adjacency of the PLD accountant (a null element contributes nothing to the sum). The shuffled runs do not model other details of the released implementations, such as reusing a single permutation (fixed batches) or the finite shuffle buffers of settings $14$, $19$ and $20$.

\para{Batches and multiple epochs.}
For the upper bound, revealing within each epoch the position of the distinguished example inside its batch together with the other $B-1$ examples in every batch reduces shuffling with $n$ batches of size $B$ to shuffling $\shuf{n}{\GaussMech{\sigma}}$ of $n$ elements (up to known shifts of the batch sums, which cancel by post-processing). We then construct the single-epoch envelope pair $(P_*,Q_*)$ of \cref{sec:pld-accounting}, combining a grid of clone probabilities together with the unamplified Gaussian pair, and compose its PLD across the $E$ epochs in both directions using the shared-grid computation of \cref{app:implementation}. The sufficient $\sigma$ values in \cref{sec:dpsgd} use this shared-grid computation with $2^{12}$ cells per octave, $1001$ logarithmically spaced clone probabilities, a step of $0.005$ in privacy loss for the sampled Gaussian hockey-stick bounds, and final loss step $10^{-4}$. At the sufficient $\sigma$ values reported in \cref{tab:dpsgd-noise}, coarsening the clone-probability grid from $1001$ to $101$ points changes the composed $\eps$ by less than $\resCloneGridDpsgd\%$ across all $20$ settings. Compared with using a single clone probability, the envelope lowers the sufficient $\sigma$ in all $20$ settings (by a median of $\resEnvelopeMedian\%$ and up to $\resEnvelopeMax\%$), reducing the maximum ratio of sufficient to necessary $\sigma$ from $\resSingleCloneSufNecMax$ to $\resSufNecMax$ and the median from $\resSingleCloneSufNecMedian$ to $\resSufNecMedian$.

For the lower bounds, we compute the necessary $\sigma$ in \cref{fig:dpsgd,tab:dpsgd-noise} with the multi-pass lower bound of \cref{sec:lower-bound,app:lower-bound} (implemented as in \cref{app:implementation}), and also plot in \cref{fig:dpsgd} the max-statistic lower bound for $E$ epochs with independent permutations \citep{CGKKMSZ24b}, computed with its released code. Both are lower bounds for the instance in which every other contribution points against the distinguished one, as in \cref{sec:lower-bound}, and like our upper bound they depend on the batch size only through $n$.

\para{Calibration.}
For Poisson subsampling, our upper bound, and the max-statistic lower bound, we find $\sigma$ by root finding in $\ln(\sigma)$ to a relative accuracy of $10^{-4}$, reporting the larger endpoint of the final bracket (at which the reported guarantee holds) for the upper bounds and the smaller endpoint (at which it is still violated) for the lower bound. For our multi-pass lower bound we search in $\ln(\sigma)$ down to a relative step of $2\times10^{-4}$ and report a directly evaluated $\sigma$ at which $\underline H(\eps)>\delta$, ruling out all smaller noise multipliers by Gaussian post-processing. The resulting noise multipliers and ratios for all $20$ settings are collected in \cref{tab:dpsgd-noise}.

\begin{table}[!htbp]
\centering\footnotesize\setlength{\tabcolsep}{2.5pt}
\begin{tabular}{@{}rl rr r rr rr rrr@{}}
\toprule
 & & & & Reported & \multicolumn{2}{c}{Poisson $\sigma$} & \multicolumn{2}{c}{Shuffling $\sigma$} & \multicolumn{3}{c}{Ratio}\\
\cmidrule(lr){6-7}\cmidrule(lr){8-9}\cmidrule(l){10-12}
ID & Ref. & $n$ & $E$ & $\sigma$ & original & matching & necessary & sufficient & \shortstack[r]{sufficient/\\matching} & \shortstack[r]{necessary/\\matching} & \shortstack[r]{sufficient/\\necessary}\\
\midrule
1 & \cite{TB21} & 29 & 40 & 2.15 & 1.81 & 1.82 & 2.39 & 3.09 & 1.70 & 1.31 & 1.29 \\
2 & \cite{TB21} & 24 & 40 & 3.32 & 2.68 & 2.72 & 3.18 & 4.14 & 1.53 & 1.17 & 1.30 \\
3 & \cite{TB21} & 48 & 30 & 1.54 & 1.32 & 1.33 & 1.96 & 2.50 & 1.88 & 1.47 & 1.27 \\
\addlinespace[3pt]
4 & \cite{BPS19} & 117 & 61 & 0.8 & 0.749 & 0.752 & 1.33 & 1.59 & 2.11 & 1.77 & 1.19 \\
5 & \cite{BPS19} & 1,875 & 30 & 0.6 & 0.546 & 0.546 & 1.06 & 1.18 & 2.15 & 1.94 & 1.11 \\
\addlinespace[3pt]
6 & \cite{BDSHSSKSB23} & 54 & 56 & 0.91 & 0.912 & 0.919 & 1.52 & 1.85 & 2.02 & 1.65 & 1.22 \\
7 & \cite{BDSHSSKSB23} & 63 & 48 & 0.85 & 0.848 & 0.851 & 1.44 & 1.74 & 2.04 & 1.69 & 1.21 \\
\addlinespace[3pt]
8 & \cite{YNBGIKKLMWYZ22} & 33 & 20 & 0.92 & 0.862 & 0.870 & 1.43 & 1.71 & 1.97 & 1.65 & 1.19 \\
9 & \cite{YNBGIKKLMWYZ22} & 52 & 20 & 0.83 & 0.778 & 0.780 & 1.34 & 1.58 & 2.02 & 1.72 & 1.18 \\
10 & \cite{YNBGIKKLMWYZ22} & 196 & 20 & 1.88 & 1.32 & 1.32 & 2.04 & 2.64 & 2.00 & 1.55 & 1.29 \\
11 & \cite{YNBGIKKLMWYZ22} & 328 & 20 & 0.6 & 0.584 & 0.584 & 1.10 & 1.24 & 2.12 & 1.88 & 1.13 \\
12 & \cite{YNBGIKKLMWYZ22} & 244 & 20 & 0.6 & 0.588 & 0.588 & 1.10 & 1.24 & 2.11 & 1.87 & 1.13 \\
\addlinespace[3pt]
13 & \cite{LZLRMS23} & 390 & 100 & 1 & 0.955 & 0.956 & 1.58 & 2.08 & 2.18 & 1.66 & 1.31 \\
14 & \cite{LZLRMS23} & 3,845 & 50 & 1 & 0.834 & 0.834 & 1.45 & 2.01 & 2.41 & 1.74 & 1.39 \\
\addlinespace[3pt]
15 & \cite{BDLS20} & 234 & 16 & 1.3 & 1.34 & 1.37 & 2.01 & 2.70 & 1.98 & 1.47 & 1.34 \\
16 & \cite{BDLS20} & 114 & 18 & 0.55 & 0.575 & 0.576 & 1.04 & 1.18 & 2.04 & 1.82 & 1.12 \\
17 & \cite{BDLS20} & 48 & 9 & 0.56 & 0.589 & 0.592 & 1.01 & 1.12 & 1.89 & 1.71 & 1.11 \\
18 & \cite{BDLS20} & 80 & 20 & 0.6 & 0.638 & 0.638 & 1.14 & 1.30 & 2.03 & 1.78 & 1.14 \\
\addlinespace[3pt]
19 & \cite{KSCGTT22} & 1,251 & 11 & 0.56 & 0.533 & 0.536 & 1.03 & 1.14 & 2.13 & 1.94 & 1.10 \\
20 & \cite{KSCGTT22} & 1,251 & 11 & 0.28 & 0.267 & 0.270 & 0.406 & 0.408 & 1.52 & 1.50 & 1.01 \\
\bottomrule
\end{tabular}

\caption{Noise multiplier $\sigma$ required for the reported $(\eps,\delta)$: number $n$ of batches per epoch and number $E$ of epochs of the shuffled run, the reported $\sigma$, the $\sigma$ required by Poisson subsampling with the original $q$ and $T$ and with the matching parameters ($q=1/n$ and $En$ steps), the necessary (\cref{sec:lower-bound,app:lower-bound}) and sufficient $\sigma$ for shuffling, and the ratios of the sufficient and the necessary $\sigma$ to the matching Poisson $\sigma$ and of the sufficient to the necessary $\sigma$, computed before rounding.}
\label{tab:dpsgd-noise}
\end{table}

\newpage
\printbibliography

\end{document}